\documentclass{article}

\usepackage{microtype}
\usepackage{graphicx}
\usepackage{booktabs} %
\usepackage{xspace}
\usepackage{amsmath,amsfonts,amssymb,amsthm}
\usepackage[dvipsnames]{xcolor}
\usepackage{diagbox}
\usepackage{pifont}
\usepackage{svg}
\usepackage{amsmath}
\usepackage{natbib}
\usepackage{enumitem}
\usepackage{physics}
\usepackage{mathtools}
\usepackage{algorithm}
\usepackage{algorithmic}
\usepackage{gensymb}
\usepackage{wrapfig}
\usepackage{subcaption}
\usepackage{multirow}
\usepackage{footmisc}
\usepackage{empheq}
\usepackage{cases}
\usepackage{adjustbox}
\usepackage{array}
\usepackage{tabu}
\usepackage{flushend}
\usepackage{multirow}
\newcolumntype{P}[1]{>{\centering\arraybackslash}p{#1}}

\DeclareMathOperator*{\argmin}{arg\,min}

\usepackage{hyperref}

\usepackage[accepted]{icml2022}

\usepackage[capitalize,noabbrev]{cleveref}

\theoremstyle{plain}

\theoremstyle{definition}

\theoremstyle{remark}

\usepackage[textsize=tiny]{todonotes}

\icmltitlerunning{ Visual Attention Emerges from Recurrent Sparse Reconstruction}

\begin{document}

\twocolumn[
\icmltitle{Visual Attention Emerges from Recurrent Sparse Reconstruction}

\begin{icmlauthorlist}
\icmlauthor{Baifeng Shi}{ucb}
\icmlauthor{Yale Song}{ms}
\icmlauthor{Neel Joshi}{ms}
\icmlauthor{Trevor Darrell}{ucb}
\icmlauthor{Xin Wang}{ms}
\end{icmlauthorlist}

\icmlaffiliation{ucb}{University of California, Berkeley}
\icmlaffiliation{ms}{Microsoft Research}

\icmlcorrespondingauthor{Baifeng Shi}{baifeng\_shi@berkeley.edu}

\icmlkeywords{Machine Learning, ICML}

\vskip 0.3in
]

\printAffiliationsAndNotice{}  %

\makeatletter
\DeclareRobustCommand\onedot{\futurelet\@let@token\@onedot}
\def\@onedot{\ifx\@let@token.\else.\null\fi\xspace}

\def\eg{\emph{e.g.,}\xspace} \def\Eg{\emph{E.g}\onedot}
\def\ie{\emph{i.e.,}\xspace} \def\Ie{\emph{I.e}\onedot}
\def\cf{\emph{c.f}\onedot} \def\Cf{\emph{C.f}\onedot}
\def\etc{\emph{etc}\onedot} \def\vs{\emph{vs}\onedot}
\def\wrt{w.r.t\onedot} \def\dof{d.o.f\onedot}
\def\etal{\emph{et al}\onedot}
\def\viz{\emph{viz}\onedot}
\makeatother

\newcommand{\cmark}{\ding{51}}%
\newcommand{\xmark}{\ding{55}}%

\newcommand{\model}{{VARS}\xspace} 
\newcommand\minisection[1]{\noindent \textbf{#1}}

\begin{abstract}

Visual attention helps achieve robust perception under noise, corruption, and distribution shifts in human vision, which are areas where modern neural networks still fall short. We present \model, Visual Attention from Recurrent Sparse reconstruction, a new attention formulation built on two prominent features of the human visual attention mechanism: \emph{recurrency} and \emph{sparsity}. Related features are grouped together via recurrent connections between neurons, with salient objects emerging via sparse regularization. \model adopts an attractor network with recurrent connections  that converges toward a stable pattern over time.
Network layers are represented as ordinary differential equations (ODEs), formulating attention as a recurrent attractor network that equivalently optimizes the sparse reconstruction of input using a dictionary of ``templates'' encoding
underlying patterns of data. We show that self-attention is a special case of \model with a single-step optimization and no sparsity constraint. \model can be readily used as a replacement for self-attention in popular vision transformers, consistently improving their robustness across various benchmarks. Code is released on GitHub (\url{https://github.com/bfshi/VARS}).

\end{abstract}

\section{Introduction}

One of the hallmarks of human visual perception is its robustness under severe noise, corruption, and distribution shifts~\cite{biederman1987recognition,bisanz2012learning}.  
Although having surpassed human performance on ImageNet~\cite{he2015delving},
convolutional neural networks (CNNs) are still far behind the human visual systems on robustness~\cite{dodge2017study,geirhos2017comparing} -- 
CNNs are vulnerable under random image corruption~\cite{hendrycks2019benchmarking}, adversarial perturbation~\cite{szegedy2013intriguing}, and distribution shifts~\cite{wang2019learning,djolonga2021robustness}.

Vision transformers~\cite{dosovitskiy2020image} have been reported to be more robust to image corruption and distribution shifts than CNNs under certain conditions~\cite{naseer2021intriguing,paul2021vision}. One hypothesis is that the self-attention module, a key component of vision transformers, helps improve robustness~\cite{paul2021vision}, achieving state-of-the-art performance on a variety of robustness benchmarks~\cite{mao2021towards}.  
Although the robustness of vision transformers still seems to be far behind human vision~\cite{hendrycks2021many,dodge2017study}, recent work suggests that attention is a key to achieving (perhaps human-level) robustness in computer vision.

The cognitive science literature has also suggested a close relationship between attention mechanisms and robustness in human vision~\cite{kar2019evidence,wyatte2012limits}.\footnote{Here we limit our focus on bottom-up attention, \ie, attention that fully depends on input and is not modulated by the high-level task. See~\citet{zhaoping2014understanding} for a review on different types of attention in the human visual system.} For example, visual attention in human vision has been shown to selectively amplify certain patterns in the input signal and repress others that are not desired or meaningful, leading to robust recognition under challenging conditions such
as occlusion~\cite{tang2018recurrent}, clutter~\cite{mnih2014recurrent,walther2005selective}, and severe
corruptions~\cite{wyatte2014early}. 
These findings motivate us to improve robustness of neural networks by designing attention inspired by the human visual attention.
However, despite the existing computational models of human visual attention (\eg \citet{zhaoping2014understanding}), their concrete instantiation in DNNs is still missing, and the connection between human visual attention and existing attention designs is also vague~\cite{sood2020interpreting}.

In this work, we introduce \model---Visual Attention from Recurrent Sparse reconstruction---a new attention formulation inspired by the \emph{recurrency} and \emph{sparsity} commonly observed in the human visual system. 
Human visual attention contains the process of grouping and selecting salient features
while repressing irrelevant signals~\cite{desimone1995neural}. One of its neural foundations is the recurrent connections between neurons in the same layer, as opposed to the feed-forward connections from lower to higher layers~\cite{stettler2002lateral,gilbert1989columnar,bosking1997orientation,li1998neural,lamme2000distinct}.
By iteratively connecting neurons, salient features with strong correlations are grouped and amplified~\cite{roelfsema2006cortical,o2013recurrent}.
Sparsity, on the other hand, also plays an important role in visual attention, where distracting information is muted by
the sparsity constraint and only the most salient parts of the input
remains~\cite{chaney2014hierarchical,wright2008robust}. Sparsity is also a natural
outcome of recurrent connections~\cite{rozell2008sparse}, and the two together can be used to
formulate visual attention.

Built upon this observation, \model shows that visual attention naturally emerges from a recurrent sparse reconstruction of input signals in deep neural
networks. We start from an ordinary differential equation (ODE) description of neural networks and adopt an attractor 
network~\cite{grossberg1987neural,zucker1989two,yen1998extraction} to describe the recurrently connected neurons that arrive at an equilibrium state over time. 
We then reformulate the computation model into an encoder-decoder style module and show that by adding inhibitory recurrent connections between the encoding neurons, the ODE is equivalent to optimizing the sparse reconstruction of the input using a learned dictionary of ``templates'' encoding underlying data patterns.  In practice, the feed-forward pathway of our attention module only involves optimizing the sparse reconstruction, which can be efficiently solved by the iterative shrinkage-thresholding algorithm~\cite{beck2009fast}.

We present multiple variants of \model by instantiating the learned dictionary in the sparse reconstruction as 
a \texttt{static} (input-independent), \texttt{dynamic} (input-dependent), or \texttt{static+dynamic} (combination of the two) set of templates. 
We show that the existing self-attention design~\cite{vaswani2017attention}, widely adopted in vision transformers, is a special case of \model with a dynamic dictionary but only using
a single step update of the ODE without sparsity constraints.  \model extends self-attention and exhibits higher robustness in practice.

We evaluate \model on five large-scale robustness benchmarks of naturally corrupted, adversarially perturbed and out-of-distribution images on ImageNet, where \model consistently outperforms previous methods. We also assess the quality of attention maps on human eye fixation and image segmentation datasets, and show that \model produces higher quality attention maps than self-attention.%

\section{Related Work}

\minisection{Recurrency in vision.}\ Recurrent connections are as ubiquitous as feed-forward connections in the human visual system~\cite{felleman1991distributed}. Various phenomena in human vision are credited to recurrency, such as visual grouping and pattern completion~\cite{roelfsema2002figure,o2013recurrent}, robust recognition and segmentation under clutter~\cite{vecera1997visual}, and even perceptual illusions~\cite{mely2018complementary}.

In deep learning, \citet{nayebi2018task} find that convolutional recurrent models can better capture neural dynamics in the primate visual system. Other work has designed recurrent modules to help networks attend to salient features such as contours~\cite{linsley2020recurrent}, to group and segregate an object from its context~\cite{kim2019disentangling,darrell1990segmentation}, or to conduct Bayesian inference of corrupted images~\cite{huang2020neural}. %
~\citet{zoran2020towards} combine an LSTM~\cite{hochreiter1997long} and self-attention to improve adversarial robustness. However, the LSTM is only used to generate the queries of self-attention and does not affect the attention mechanism itself.
Instead, we show visual attention emerges from recurrency and can be formulated as a recurrent contractor network, which is equivalent to optimizing sparse reconstruction of the input signals. 

\minisection{Sparsity in vision.}\ It has long been hypothesized that the primary visual cortex (V1) encodes incoming stimuli in a sparse manner~\cite{olshausen1997sparse}. \citet{olshausen1996emergence} show that localized and oriented filters resembling the simple cells in the visual
cortex can spontaneously emerge through dictionary learning via sparse coding. Some work has extended this hypothesis to the prestriate cortex such as V2~\cite{lee2007sparse}. ~\citet{rozell2008sparse} propose Locally Competitive Algorithms (LCA) as a
biologically-plausible neural mechanism for computing sparse representations in the visual cortex based on neural recurrency. 

Recently there are studies on designing sparse neural networks, but they mostly focus on reducing the computational complexity via sparse weights or connections~\cite{hoefler2021sparsity,liu2019improving,guo2018sparse}. Other work has exploited sparse data structure in neural networks ~\cite{wu2019deep,fan2020neural}. In this work, we draw a connection between sparsity and recurrency as well as visual attention and build a new attention formulation on top of it to improve the robustness of deep neural networks.

\begin{figure*}[t]
\vskip 0.2in
\begin{center}
\centerline{\includegraphics[width=2.1\columnwidth]{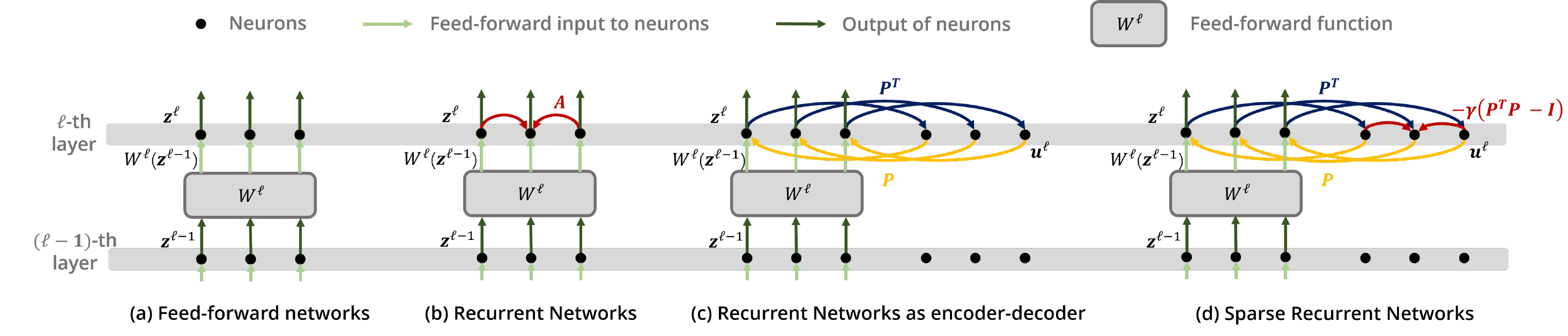}}
\caption{\textbf{Illustrations on neural dynamics.} \textit{(a)} \emph{Feed-forward networks}. The output $\mathbf{z}^{\ell-1}$ from the $(\ell-1)$-th layer's neurons is processed by the feed-forward function $W^\ell$ into $W^\ell (\mathbf{z}^{\ell-1})$, used as the input to the $\ell$-th layer's neurons. The $\ell$-th layer's neuron output $\mathbf{z}^\ell$ is identical to the input. \textit{(b)} \emph{Recurrent networks}. When the $\ell$-th layer's neurons are recurrently connected, the output $\mathbf{z}^\ell$ is wired back by weight matrix $\mathbf{A}$ serving as an additional input to the neurons. The final output is the steady state $\mathbf{z}^{\ell^\ast} = \mathbf{A} \mathbf{z}^{\ell^\ast} + W^\ell (\mathbf{z}^{\ell-1})$. \textit{(c)} \emph{Recurrent networks as encoder-decoder}. The neurons with recurrent connections in (b) have the same steady state as an encoder-decoder structure, where the auxiliary layer encodes $\mathbf{z}^\ell$ by $\mathbf{P}^T$ and its output is decoded by $\mathbf{P}$ and sent back. \textit{(d)} \emph{Sparse recurrent networks.} We adopt a sparse structure in the encoding by adding inhibitive recurrent connections $-\gamma(\mathbf{P}^T\mathbf{P} - \mathbf{I})$ between $\mathbf{u}^\ell$.}
\label{fig:feedforward_and_recurrent2}
\end{center}
\vskip -0.2in
\end{figure*}

\minisection{Visual attention and robustness.}\ Attention widely exists in human brains~\cite{scholl2001objects} and is adopted in machine learning models~\cite{vaswani2017attention}.
A number of studies focus on object-based and bottom-up attention. 
Various computational models have been proposed for visual attention~\cite{koch1987shifts,itti2001computational,zhaoping2014understanding}, where
neural recurrency help highlight salient features. 

In deep learning, the attention mechanism is widely used to process language or
visual data~\cite{vaswani2017attention,dosovitskiy2020image}. The recently proposed self-attention based vision transformers~\cite{dosovitskiy2020image} can scale to large datasets better than conventional CNNs and achieve better robustness under distribution shifts~\cite{mao2021towards,naseer2021intriguing,paul2021vision}.  

In a complementary line of work to address model robustness, researchers have shown that model robustness can be 
improved through data augmentation~\cite{geirhos2018imagenet,rebuffi2021data}, designing more
robust model architectures~\cite{dong2020adversarially} and training strategies~\cite{madry2017towards,wang2021robust}. Some other work~\cite{machiraju2021bio,shi2020informative} improves model robustness from a bio-inspired view. In this work, we propose a new attention design, (partially) inspired by human vision systems,  which can be used as a replacement for self-attention in vision transformers to further improve model robustness.

\section{VARS Formulation}

In this section, we first formulate the neural recurrency based on an
ODE description of neural dynamics (Section~\ref{sec:recurrency}) and show its
equivalence to optimizing the sparse reconstruction of the inputs
(Section~\ref{sec:sparse_recon}). 
Then we draw a connection between visual attention and recurrent sparse reconstruction and describe the design of \model (Section~\ref{sec:attention}).
We find self-attention is a special case of \model (Section~\ref{sec:compare_with_self_att}) and give different model instantiations in Section~\ref{sec:instantiation}.

\subsection{Neural Recurrency}
\label{sec:recurrency}

\minisection{ODE descriptions of neural dynamics.}
We start with an ODE description of feed-forward neural networks~\cite{dayan2001theoretical}. Let  $\mathbf{z}^\ell \in \mathbb{R}^d$ denote the output of the $\ell$-th layer's neurons in a neural network.\footnote{Although $\mathbf{z}^\ell$ may have shape like $h \times w \times c$, in our formulation we always use the vectorized version of the input, \ie $d = hwc$.\label{footnote}}
In a feed-forward neural network, the output of the $\ell$-th layer is 
\begin{equation}\small
   \mathbf{z}^\ell = W^\ell (\mathbf{z}^{\ell-1}),
   \label{eq:feedforward}
\end{equation}
where $W^\ell(\cdot)$ is a feed-forward function (\eg a convolutional or fully connected operator). This equation can be viewed as an \emph{equilibrium state}  ($\dv{\mathbf{z}^\ell}{t} = 0$) of the following differential equation:
\begin{equation}\small
\label{eq:ode_wo_rec}
    \dv{\mathbf{z}^\ell}{t} = -\mathbf{z}^\ell + W^\ell (\mathbf{z}^{\ell-1}),
\end{equation} 
which defines the neural dynamics of $\ell$-th layer's neurons. This can be seen as the (simplified) dynamics of biological neurons: $\mathbf{z}^\ell$ is the membrane potential which is charged by the feed-forward input $W^\ell (\mathbf{z}^{\ell-1})$ and discharged by the self leakage $-\mathbf{z}^\ell$~\cite{dayan2001theoretical}. Note that in the feed-forward case, the output $\mathbf{z}^\ell$ of the neurons is identical to the input $W^\ell (\mathbf{z}^{\ell-1})$. Figure~\ref{fig:feedforward_and_recurrent2}(a) provides an illustration.

\minisection{Horizontal recurrent connections.} In the feed-forward case, the input $W^\ell (\mathbf{z}^{\ell-1})$ to the $\ell$-th layer solely depends on the output $\mathbf{z}^{\ell-1}$ from the previous layer. However, when the neurons in the $\ell$-th layer are recurrently connected (illustrated in Figure~\ref{fig:feedforward_and_recurrent2}(b)),   
the input also depends on the $\ell$-th layer's output $\mathbf{z}^\ell$ itself, which we denote by $\widetilde{W}^\ell (\mathbf{z}^{\ell-1}, \mathbf{z}^\ell)$. Therefore, the output of a recurrently connected layer is the equilibrium state of an updated differential equation 
\begin{equation}\small
    \dv{\mathbf{z}^\ell}{t} = -\mathbf{z}^\ell + \widetilde{W}^\ell (\mathbf{z}^{\ell-1}, \mathbf{z}^\ell).
    \label{eq:equilibrium}
\end{equation}
In this case, the equilibrium $\mathbf{z}^{\ell^\ast} = \widetilde{W}^\ell (\mathbf{z}^{\ell-1}, \mathbf{z}^{\ell^\ast})$\footnote{We use asterisks to denote equilibrium states.} typically does not have a closed-form
solution, and in practice the differential equation is often solved by rolling out
each step of updates as in recurrent neural networks (RNNs) (\eg~\citet{hochreiter1997long}) or by using root-finding
techniques~\cite{bai2019deep}. 

Following the previous work by \citet{zhaoping2014understanding}, we decompose $\widetilde{W}^\ell (\mathbf{z}^{\ell-1}, \mathbf{z}^\ell)$ into a feed-forward input $W^\ell (\mathbf{z}^{\ell-1})$ and an additional recurrent input $\mathbf{A}^\ell \mathbf{z}^\ell$ and rewrite Equation~\ref{eq:equilibrium} as 
\begin{equation}\small
\label{eq:ode}
    \dv{\mathbf{z}^\ell}{t} = -\mathbf{z}^\ell + \mathbf{A}^\ell \mathbf{z}^\ell + \mathbf{x}^\ell
\end{equation}
where $\mathbf{x}^\ell = W^\ell (\mathbf{z}^{\ell-1})$ is the feed-forward input of the neurons in the $\ell$-th layer and $\mathbf{A}^\ell \in \mathbb{R}^{d \times d}$, often assumed symmetric and positive semi-definite~\cite{hopfield1984neurons,cohen1983absolute}, 
is the weight of horizontal recurrent connections of neurons in the $\ell$-th layer (Figure~\ref{fig:feedforward_and_recurrent2}(b)). 
Note that Equation~\ref{eq:ode} is also a special case of the continuous attractor neural network~\cite{wu2016continuous}, which is used as a computational model for various neural behaviors such as visual attention~\cite{zhaoping2014understanding}. In what follows, for simplicity and without loss of generality, we only focus on a single-layer scenario and omit the superscript $\ell$.

\subsection{Recurrency Entails Sparse Reconstruction}
\label{sec:sparse_recon}
We have defined the neural recurrency in ODEs and now we build its connection to the optimization of sparse reconstruction to understand the functionality of recurrency. 

We notice that Equation~\ref{eq:ode} can also be viewed as an encoder-decoder structure. Since $\{\mathbf{A} \ | \ \mathbf{A} \in \mathcal{S}^{d}_+ \} = \{ \mathbf{P} \mathbf{P}^T \ | \ \mathbf{P} \in \mathbb{R}^{d \times d^\prime}, d^\prime \ge d \}$, we can reparameterize $\mathbf{A}$ as $\mathbf{P}\mathbf{P}^T$, and turn Equation~\ref{eq:ode} into
\begin{equation}\small
\label{eq:conv-deconv}
    \dv{\mathbf{z}}{t} = -\mathbf{z} + \mathbf{P}\mathbf{P}^T\mathbf{z} + \mathbf{x}, 
\end{equation}
which has the same steady state solution as
\vspace{-0.5em}
\begin{small}
\begin{empheq}[left=\empheqlbrace]{align}
    &\dv{\mathbf{z}}{t} = -\mathbf{z} + \mathbf{P}\mathbf{u} + \mathbf{x}, \\
    &\dv{\mathbf{u}}{t} = -\mathbf{u} + \mathbf{P}^T\mathbf{z}.
\end{empheq}
\end{small}\noindent
Here, we introduce an auxiliary layer $\mathbf{u} \in \mathbb{R}^{d^\prime}$ that receives input $\mathbf{P}^T\mathbf{z}$ and converges to $\mathbf{u}^\ast = \mathbf{P}^T\mathbf{z}^\ast$. Meanwhile $\mathbf{z}$ converges to $\mathbf{z}^\ast = \mathbf{P}\mathbf{u}^\ast + \mathbf{x}$. We can view the steady state solution as an equilibrium between an encoder and a decoder, where the column vectors of
$\mathbf{P}$ serve as atoms (or ``templates'') of a dictionary,  $\mathbf{u}^\ast$ is the encoding of $\mathbf{z}^\ast$ through the template matching
$\mathbf{P}^T\mathbf{z}^\ast$, and $\mathbf{z}^\ast = \mathbf{P}\mathbf{u}^\ast + \mathbf{x}$ is the decoding of $\mathbf{u}^\ast$ plus a residual connection
$\mathbf{x}$. (Figure~\ref{fig:feedforward_and_recurrent2}(c)).

Next, we show that this encoder-decoder structure naturally connects with the hypothesis of sparse coding in V1, which states that the encoding of visual signals should be sparse~\cite{olshausen1997sparse}. 
Moreover, sparsity naturally emerges as we build the inhibitive recurrent connections between the encoding neurons ($\mathbf{u}$ in our case)~\cite{rozell2008sparse}. 
To show this, we follow~\citet{rozell2008sparse} and add recurrent connections between $\mathbf{u}$, modeled by the weight matrix $- \gamma (\mathbf{P}^T\mathbf{P} - \mathbf{I})$. 
In addition, we add hyperparameters $\alpha$ and $\beta$ to
control the strength of self-leakage, and the element-wise activation functions $\mathit{g}(\cdot)$ to gate the output from the neurons~\cite{dayan2001theoretical}. As a result, we update the dynamics of $\mathbf{z}$ and $\mathbf{u}$ as (see also Figure~\ref{fig:feedforward_and_recurrent2}(d)):

\vspace{-0.8em}
\begin{small}
\begin{empheq}[left=\empheqlbrace]{align}
    &\dv{\mathbf{z}}{t} = -\alpha \mathbf{z} + \mathbf{P}\mathit{g}(\mathbf{u}) + \mathbf{x}, \label{eq:fixed_dynamics_1}\\
    &\dv{\mathbf{u}}{t} = -\beta \mathbf{u} - \gamma (\mathbf{P}^T\mathbf{P} - \mathbf{I}) \mathit{g}(\mathbf{u}) + \mathbf{P}^T \mathbf{z} \label{eq:fixed_dynamics_2}.
\end{empheq}
\end{small}\noindent
By taking $\alpha = 1$ and $\beta = \gamma = 2$ and choosing $\mathit{g}$ as an element-wise thresholding function $\mathit{g}(\mathbf{u}_i) = \mathit{sgn}(\mathbf{u}_i) \cdot (\lvert \mathbf{u}_i \rvert - \frac{\lambda}{2})_+$, where $sgn(\cdot)$ is the sign function, $(\cdot)_+$ is ReLU, and $\lambda$ controls the sparse constraint (see Appendix for more details), Equation \ref{eq:fixed_dynamics_1}-\ref{eq:fixed_dynamics_2} have the equilibrium state as

\vspace{-0.8em}
\begin{small}
\begin{empheq}[left=\empheqlbrace]{align}
\label{eq:static_a}
    \widetilde{\mathbf{u}}^\ast &= \argmin_{\widetilde{\mathbf{u}} \in \mathbb{R}^{d^\prime}} \frac{1}{2} ||\mathbf{P} \widetilde{\mathbf{u}} - \mathbf{x}||^2_2 + \lambda ||\widetilde{\mathbf{u}}||_1, \\
\label{eq:static_b}
    \mathbf{z}^\ast &= \mathbf{P} \widetilde{\mathbf{u}}^\ast + \mathbf{x},
\end{empheq}
\end{small}\noindent
where $\widetilde{\mathbf{u}} = \mathit{g}(\mathbf{u})$ is the gated output of the encoding neurons. We can see that by adding the recurrent connections in $\mathbf{u}$ and the activation functions $\mathit{g}(\cdot)$, the output of $\ell$-th layer's neurons is not only a simple ``copy-paste'' of the feed-forward input $\mathbf{x}$ (Equation \ref{eq:feedforward}) but also with a sparse reconstruction of the input $\mathbf{P} \widetilde{\mathbf{u}}^\ast$. This formulation also indicates that solving the dynamics of a sparse recurrent network is equivalent to sparse reconstruction of the input signal.

\subsection{\model: Attention from Sparse Reconstruction}
\label{sec:attention}
The core design of \model is based on the observation that visual attention is achieved through (i) \emph{grouping}
different features and different locations into separate objects and (ii)
\emph{selecting} the most salient objects and suppressing distracting or noisy ones~\cite{desimone1995neural}. Note that the dynamics of sparse encoding $\mathbf{u}$ (Equation \ref{eq:fixed_dynamics_2}) contains a similar process, where the features in $\mathbf{z}$ are grouped by each template $\mathbf{P}^\mu$ through $(\mathbf{P}^\mu)^T \mathbf{z}$ and fed into the encoding $\mathbf{u}_\mu$,\footnote{We denote the $\mu$-th column of $\mathbf{P}$ by $\mathbf{P}^\mu$. For example, in the binary case, if each template contains an object, \ie $\mathbf{P}^\mu_i$ = 1 if the object occupies the location $i$, then $(\mathbf{P}^\mu)^T \mathbf{x} = \sum_{i \in \{i | \mathbf{P}^\mu_i = 1\}} \mathbf{x}_i$ is the collection of all features in locations that the object occupies. } 
meanwhile the recurrent term $- \gamma (\mathbf{P}^T\mathbf{P} - \mathbf{I}) \mathit{g}(\mathbf{u})$ imposes a sparse structure on $\mathbf{u}$ so that only the $\mathbf{u}_\mu$ that encodes the most salient template objects will survive. 

Here we introduce VARS, a module that achieves visual attention via sparse reconstruction following the formulation in Equation \ref{eq:static_a}-\ref{eq:static_b}, \ie \model takes a feed-forward input $\mathbf{x}$ and outputs the sparse reconstruction $\mathbf{P}\widetilde{\mathbf{u}}^\ast$ of $\mathbf{x}$ and a residual term. The \model module can be plugged into neural networks to help attend features. 

\begin{figure}[t]
\vskip 0.2in
\centering
\centerline{\includegraphics[width=1\columnwidth]{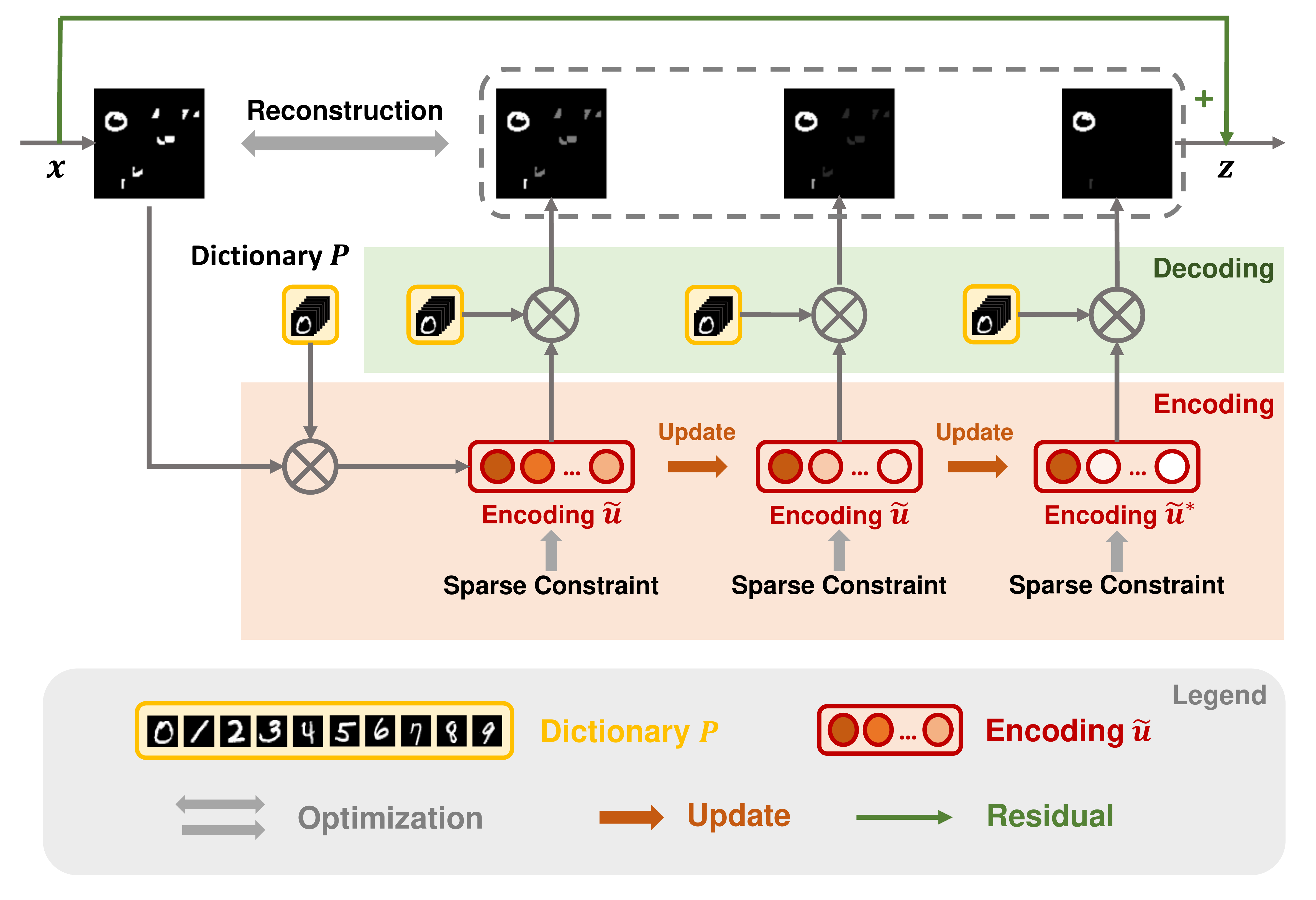}}
\caption{\textbf{Overview of \model}. First, we initialize $\widetilde{\mathbf{u}}$ as $\mathbf{P}^T \mathbf{x}$, the encoding of the input. Then, for each iteration, we update $\widetilde{\mathbf{u}}$ to minimize the reconstruction error between $\mathbf{x}$ and the decoded $\mathbf{P} \widetilde{\mathbf{u}}$, as well as the sparsity constraint. After multiple steps, the converged $\widetilde{\mathbf{u}}^\ast$ is decoded and output together with a residual term.}
\label{fig:encoder_decoder}
\vskip -0.2in
\end{figure}

In practice, \model optimizes the sparse reconstruction (Equation \ref{eq:static_a}) iteratively, as illustrated in Figure \ref{fig:encoder_decoder}. First, we initialize $\widetilde{\mathbf{u}}$ as the encoding of input $\mathbf{x}$, \ie $\widetilde{\mathbf{u}} \leftarrow \mathbf{P}^T \mathbf{x}$ (the red block in Figure \ref{fig:encoder_decoder}). Then, for each iteration, we decode $\widetilde{\mathbf{u}}$ into $\mathbf{P} \widetilde{\mathbf{u}}$ (the green block in Figure \ref{fig:encoder_decoder}) and update $\widetilde{\mathbf{u}}$ by minimizing the reconstruction error $\frac{1}{2} ||\mathbf{P} \widetilde{\mathbf{u}} - \mathbf{x}||^2_2$ and the sparsity constraint $\lambda ||\widetilde{\mathbf{u}}||_1$. We adopt the update rule in the Iterative Shrinkage Thresholding Algorithm (ISTA)~\cite{beck2009fast}, \ie each update is
\begin{equation}\small
\label{eq:ista}
    \widetilde{\mathbf{u}} \leftarrow \mathcal{S}_{\frac{\lambda}{L}} \left( \widetilde{\mathbf{u}} - \frac{1}{L}(\mathbf{P}^T\mathbf{P} \widetilde{\mathbf{u}} - \mathbf{P}^T \mathbf{x}) \right),
\end{equation}
where $\mathcal{S}_\lambda(x) = sgn(x) \cdot (\lvert x \rvert - \lambda)_+$ is an element-wise thresholding function, and $L$ is the largest singular value of $\mathbf{P}^T\mathbf{P}$.
After multiple updates, we decode the converged $\widetilde{\mathbf{u}}^\ast$ into $\mathbf{P} \widetilde{\mathbf{u}}^\ast$ and output it together with a residual term.

Overall, VARS groups features of different objects in the input into different encoding neurons, and the most salient objects (\eg ``0'' in the input in Figure \ref{fig:encoder_decoder}) are preserved while other distractors are suppressed by sparse reconstruction. \model can be easily plugged into any neural network and is computationally efficient thanks to the ISTA algorithm with fast convergence (see Section \ref{sec:exp}).

\subsection{Self-Attention as a Special Case of \model}
\label{sec:compare_with_self_att}
Here we show that self-attention can be viewed as a special case of \model using a dynamic instantiation of the dictionary with single step approximation and no sparsity constraint.

\minisection{Self-attention formulation.} In self-attention, the input $\mathbf{X} \in \mathbb{R}^{N \times C}$ contains $N$ tokens and $C$ channels. We use the superscript $\mu$ for the channel index and the subscript $i$ for the token index, \eg $\mathbf{X}^\mu_i$ is the $\mu$-th channel in the $i$-th token. In each head, self-attention gives the output $\mathbf{Z}$ by
\begin{equation}\small
\label{eq:self-att}
\mathbf{Z}^\mu_i = \sum_j \mathbf{K}(\mathbf{X}, \mathbf{X})_{ij} \cdot \mathbf{X}^\mu_j + \mathbf{X}^\mu_i, 
\end{equation}
where $\mathbf{K}(\mathbf{X}, \mathbf{X}) \in \mathbb{R}^{N \times N} $ measures the similarity between tokens in $\mathbf{X}$, \ie $\mathbf{K}(\mathbf{X}, \mathbf{X})_{ij} = e^{(\mathbf{W}\mathbf{X}_i)^T (\mathbf{W}^\prime \mathbf{X}_j)}$, with $\mathbf{W}$ and $\mathbf{W}^\prime$ as query and key
projections.\footnote{Here we ignore the value projection (as in non-local blocks~\cite{wang2018non}) as well as the normalization term.} Self-attention is compute intensive given its quadratic computational complexity. 
Performer~\cite{choromanski2020rethinking}, a recently proposed variant of vision transformers, approximates the similarity kernel with the inner product of the feature maps, \ie $\mathbf{K}(\mathbf{X}, \mathbf{X}) \approx \Phi(\mathbf{X}) \Phi^\prime(\mathbf{X})^T$, where $\Phi(\mathbf{X}), \Phi^\prime(\mathbf{X}) \in \mathbf{R}^{N \times C^\prime}$ are specific (random) feature maps of $\mathbf{X}$. 

\minisection{Connection between self-attention and \model.} Following the formulation in Performer, we can rewrite Equation~\ref{eq:self-att} as 
\begin{equation}\small
\label{eq:self-att_linear}
\mathbf{Z}^\mu = \Phi(\mathbf{X}) \Phi(\mathbf{X})^T \mathbf{X}^\mu + \mathbf{X}^\mu.
\end{equation}
Here we use a symmetric similarity kernel by setting $\mathbf{W} = \mathbf{W}^\prime$, which means $\Phi = \Phi^\prime$. This feed-forward computation is a single-step Euler update\footnote{with initialization $\mathbf{Z}^\mu = \mathbf{X}^\mu$ and step-size of 1.} of the differential equation
\begin{equation}\small
\label{eq:ode_self_att}
    \dv{\mathbf{Z}^\mu}{t} = -\mathbf{Z}^\mu + \Phi(\mathbf{X}) \Phi(\mathbf{X})^T \mathbf{Z}^\mu + \mathbf{X}^\mu,
\end{equation}
which has a similar form with the ODE description of a recurrent layer (Equation~\ref{eq:conv-deconv}), except that Equation~\ref{eq:ode_self_att} uses $\Phi(\mathbf{X})$ as the dictionary which is dependent on the specific input $\mathbf{X}$ while Equation~\ref{eq:conv-deconv} uses a static dictionary $\mathbf{P}$ learned from the entire dataset. This shows self-attention is a variant of recurrent networks using a dynamic dictionary. See Appendix~\ref{apdx:sec:dict_vis} for the visualization of the dynamic dictionary.

However, compared to \model (Equation \ref{eq:static_a}-\ref{eq:static_b}), self-attention (Equation \ref{eq:ode_self_att}) does not have the inhibitive recurrent connections (Equation \ref{eq:fixed_dynamics_2}) to turn into sparse reconstruction, and updates the ODE (Equation \ref{eq:ode_self_att}) only via a single step. Therefore, we introduce \model with a dynamic dictionary:

\vspace{-1.3em}
\begin{small}
\begin{empheq}[left=\empheqlbrace]{align}
\label{eq:dynamic_a}
    \widetilde{\mathbf{U}}^{\mu^\ast} &= \argmin_{\widetilde{\mathbf{U}}^{\mu}} \frac{1}{2} ||\Phi(\mathbf{X}) \widetilde{\mathbf{U}}^{\mu} - \mathbf{X}^{\mu}||^2_2 + 2\lambda ||\widetilde{\mathbf{U}}^\mu||_1 \\
\label{eq:dynamic_b}
    \mathbf{Z}^{\mu^{\ast}} &= \Phi(\mathbf{X})\widetilde{\mathbf{U}}^{\mu^{\ast}} + \mathbf{X}^{\mu},
\end{empheq}
\end{small}\noindent
which optimizes the sparse reconstruction of each channel $\mathbf{X}^\mu$ in the input. We refer \model with a static dictionary to as \texttt{\model-S} (Equation \ref{eq:static_a}-\ref{eq:static_b}) and \model with an dynamic dictionary as \texttt{\model-D} (Equation \ref{eq:dynamic_a}-\ref{eq:dynamic_b}). %

\subsection{\model Instantiations}
\label{sec:instantiation}
\minisection{Dictionary designs.} So far, we introduced two instantiations of \model: \model with a static dictionary $\mathbf{P}$ (\texttt{\model-S}) %
(Equation~\ref{eq:static_a}-\ref{eq:static_b}) and \model with a dynamic dictionary $\Phi(\mathbf{X})$ (\texttt{\model-D}) %
(Equation~\ref{eq:dynamic_a}-\ref{eq:dynamic_b}). Both
variants have their own merits as $\mathbf{P}$ learns a general pattern of the dataset while $\Phi(\mathbf{X})$  captures the specialized information on a per input basis. Therefore, we consider a third instantiation of \model,
\texttt{\model-SD} to combine both the static and dynamic dictionaries as $[\mathbf{P}; \Phi(\mathbf{X})]$, \ie using the union of atoms in $\mathbf{P}$ and $\Phi(\mathbf{X})$. We test all three variants in our experiments.

\textbf{Instantiation of $\mathbf{P}$}. In theory, $\mathbf{P}$ can be any real
matrix of the specific shape. However, since each column of $\mathbf{P}$ is a template which is a signal in the spatial feature domain, we may impose certain
inductive biases such as translational symmetry when instantiating $\mathbf{P}$. To this end, we design $\mathbf{P}$ by making its templates kernels with different translations, which means $\mathbf{P}^T$ is a convolution layer, \ie
$(\mathbf{P}^T\mathbf{x})_\uparrow = \mathit{conv}(\mathbf{x}_\uparrow)$ where $(\cdot)_\uparrow$ unflattens a vector into a 2D signal. Then $\mathbf{P}$ is a deconvolution layer with its kernel shared with the convolution layer, used in both \texttt{static} and \texttt{static+dynamic} dictionaries.

\begin{table}[t]
    \centering
    \caption{\textbf{Dataset overview.} We evaluate \model on five robustness benchmarks and perform evaluation in three additional settings.\vspace{1mm}}
    \adjustbox{width=\linewidth}{
    \begin{tabular}{cll}
    \toprule
          & Dataset Name & Type \\
         \midrule\midrule
         \multirow{5}{*}{\rotatebox[origin=c]{90}{Robustness}}& ImageNet-C (IN-C)~\cite{hendrycks2019benchmarking} & Natural corruption \\
         &ImageNet-R (IN-R)~\cite{hendrycks2021many} & Out of distribution\\
         &ImageNet-SK (IN-SK)~\cite{wang2019learning} & Out of distribution\\
         &PGD~\cite{madry2017towards} & Adversarial attack \\
         &ImageNet-A (IN-A)~\cite{hendrycks2021natural} & Natural adv. example \\
         \midrule
        \multirow{3}{*}{\rotatebox[origin=c]{90}{Others}} & PACS~\cite{li2017deeper} & Domain generalization \\
        & PASCAL VOC~\cite{everingham2010pascal} & Semantic segmentation \\
        & MIT1003~\cite{judd2009learning} & Human eye fixation \\
        \bottomrule
    \end{tabular}}
    \label{tab:data}
    \vskip -0.3in
\end{table}

\section{Experiments}
\label{sec:exp}
We test \model on five robustness benchmarks on the ImageNet dataset including naturally  corrupted, out of distribution, and adversarial images (Section~\ref{sec:imagenet}). We also evaluate our models on the following settings: domain generalization (Section \ref{sec:exp_dg}), image segmentation, and human eye fixation predictions (Section \ref{sec:exp_seg_eye}). Finally, we analyze and ablate the design choices of \model in Section~\ref{sec:ablation}.

\minisection{Experimental Setup.} We evaluate on multiple datasets (Table~\ref{tab:data}) and the models are pretrained on ImageNet-1K~\cite{deng2009imagenet}. For baselines, we consider DeiT~\cite{touvron2021training}, a commonly-used vision transformer model, and RVT~\cite{mao2021towards}, the state-of-the-art vision transformer on various robustness benchmarks. For generality, we also test on GFNet~\cite{rao2021global} using a linear token-mixer instead of self-attention. We apply \model in the baselines by replacing the global operators (self-attention or token mixer). For all the models, we adopt the convolutional patch-embedding ~\cite{xiao2021early} to facilitate training and also apply Performer approximation which \model is also built on. We use $\ast$ to denote the modified baselines. Results of both original and modified baselines are reported.%

\begin{figure}[t]
\centering
\centerline{\includegraphics[width=0.9\columnwidth]{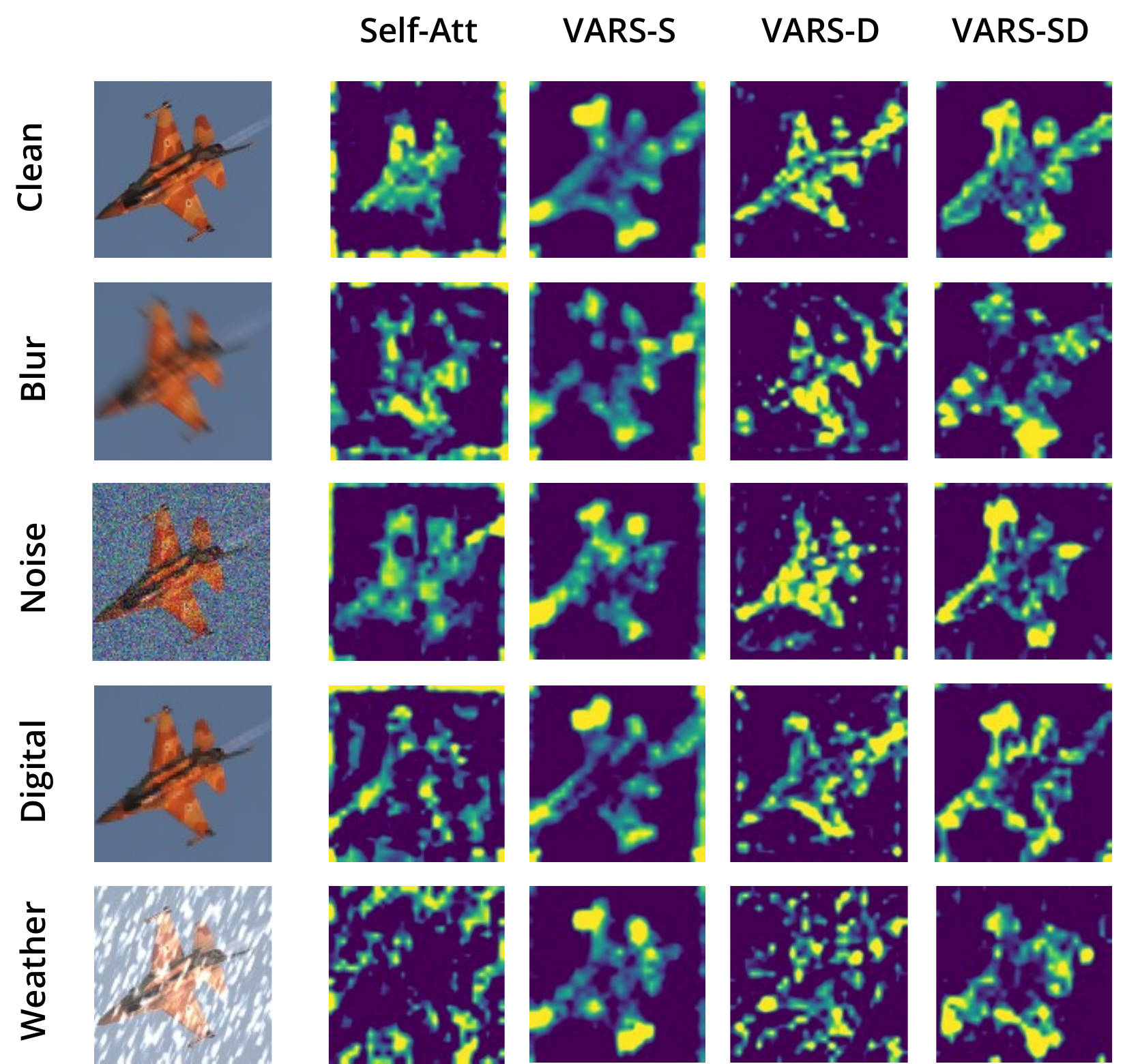}}
\vskip -0.1in
\caption{\textbf{Attention maps under image corruption}. We can see \model consistently highlights the core parts of the object (airplane) while self-attention can miss them (\eg weather). The attention maps of \model-SD are most stable and sharp. }
\label{fig:att_robust}
\vskip -0.2in
\end{figure}

\begin{table*}[t]
\caption{\textbf{Evaluation results on robustness benchmarks.} We find that \model consistently improves over the self-attention counterparts (DeiT$^*$, GFNet$^*$ and RVT$^*$). \texttt{\model-SD} outperforms or is on par with previous methods despite using a weaker initial model. The best performance of each vision transformer architecture is bold and the underlined values are the overall state-of-the-art performance.\vspace{1mm}}
\label{table:ablation_imagenet}
\centering
\adjustbox{width=\linewidth}{
\begin{tabular}{clP{0.1\textwidth}P{0.1\textwidth}P{0.1\textwidth}P{0.1\textwidth}P{0.1\textwidth}P{0.1\textwidth}P{0.1\textwidth}P{0.1\textwidth}}
\toprule
& Model  & GFLOPs & Params.(M) & Clean$\uparrow$ & IN-C$\downarrow$ & IN-R$\uparrow$ & IN-SK$\uparrow$  & PGD$\uparrow$  & IN-A$\uparrow$ \\
\midrule
\midrule
{\multirow{4}{*}{\rotatebox[origin=c]{90}{CNNs}}} &{RegNetY-4GF}~\cite{radosavovic2020designing} & 4.0 & 20.6 & 79.2 & 68.7 & 38.8 & 25.9 & 2.4 & 8.9  \\
&{ResNet50}~\cite{he2016deep} & 4.1 & 25.6 & 79.0 & 65.5 & 42.5 & 31.5 & 12.5 & 5.9  \\
&{ResNeXt50-32x4d}~\cite{xie2017aggregated} & 4.3 & 25.0 & 79.8 & 64.7 & 41.5 & 29.3 & 13.5 & 10.7\\
&{InceptionV3}~\cite{szegedy2016rethinking} &  5.7 & 27.2 & 77.4 & 80.6 & 38.9 & 27.6 & 3.1 & 10.0  \\
\midrule
{\multirow{18}{*}{\rotatebox[origin=c]{90}{Transformers}}} &PiT-Ti~\cite{heo2021rethinking} & 0.7 & 4.9 &  72.9 & 69.1 & 34.6 & 21.6 & 5.1 & 6.2 \\
&ConViT~\cite{d2021convit} & 1.4  & 5.7 & 73.3 & 68.4 & 35.2 & 22.4 & 7.5 & 8.9 \\
&PVT~\cite{wang2021pyramid} & 1.9 & 13.2 & 75.0 & 79.6 & 33.9 & 21.5 & 0.5 & 7.9 \\
&DeiT~\cite{touvron2021training} & 1.3 & 5.7 & 72.2 & 71.1 & 32.6 & 20.2 & 6.2 & 7.3  \\
&GFNet~\cite{rao2021global}  & 1.3 & 7.5 & 74.6 & 65.9 & 40.4 & 27.0 & 7.6 & 6.3 \\
&RVT~\cite{mao2021towards} & 1.3 & 8.6 & 78.4 & 58.2 & \underline{43.7} & 30.0 & 11.7 & 13.3 \\
\cmidrule{2-10}
\cmidrule{2-10}
&DeiT$^\ast$ & 1.3 & 5.7 &  74.7 & 67.0 & 34.5 & 21.7 & 11.9 & 9.4  \\
&w/ \model-S & 1.0 & 6.8 &  73.7 & 69.8 & 36.8 & 24.8 & 10.8 & 4.9 \\
&w/ \model-D & 1.4 & 5.4 &  75.6 & 64.9 & 39.6 & \textbf{27.5} & \textbf{13.7} & 10.2 \\
&w/ \model-SD & 1.4 & 5.8 & \textbf{76.5} & \textbf{62.5} & \textbf{40.2} & \textbf{27.5} & 13.4 & \textbf{11.5}\\
\cmidrule{2-10}
&GFNet-Ti$^\ast$ & 1.3 & 7.5 & 74.6 & 65.9 & 40.4 & 27.0 & 7.6 & 6.3 \\
&w/ \model-S & 1.3 & 7.5& 74.1 & 63.5 & 40.8 & 28.6 & 9.5 & 5.8\\
&w/ \model-D & 1.9 & 9.8 &  77.8 & 58.6 & \textbf{41.2} & 29.0 & 15.9 & 12.6 \\
&w/ \model-SD & 1.9 & 10.4 & \textbf{78.2} & \underline{\textbf{57.4}} & 41.0 & \textbf{29.5} & \underline{\textbf{16.2}} & \textbf{13.0} \\
\cmidrule{2-10}
&RVT$^\ast$ & 1.3 & 8.6 & 77.6 & 60.4 & 41.7 & 28.7 & 11.1 & 11.1 \\
&w/ \model-S & 1.0 & 9.2 &  76.8 & 61.8 & \textbf{43.2} & 30.1 & 7.6 & 9.1 \\
&w/ \model-D & 1.2 & 8.0 &  78.2 & 58.7 & 42.0 & 29.8 & \textbf{11.7} & 12.4 \\
&w/ \model-SD & 1.5 & 9.2 &  \textbf{78.4} & \textbf{58.3} & 42.5 & \underline{\textbf{30.5}} & 11.4 & \underline{\textbf{13.4}}\\
\bottomrule
\end{tabular}}
\label{tab:benchmark}
\vskip -0.2in
\end{table*}

\subsection{Evaluation on Robustness Benchmarks}
\label{sec:imagenet}

We show the evaluation results on the robustness benchmarks in Table~\ref{tab:benchmark}. First, we can see vision transformers are
generally more robust than the CNN counterparts even with an order of magnitude
smaller numbers of parameters and FLOPs.
 We can also observe that \model consistently improves the baselines across
different benchmarks. For example, compared to DeiT, \texttt{\model-SD}
reduces the error rate from 67\% to 62.5\% on IN-C and improves the accuracy from 34.5\% to 40.2\% on IN-R, from 21.7\% to 27.5\% on IN-SK, which are over 5 absolute points improvements.  Similar results are observed with GFNet and RVT.

Moreover, when built on top of the RVT network design, \texttt{\model-SD} outperforms or is on par with the previous methods across the five
benchmarks. Note that \model is built upon RVT$^*$, the modified version of RVT (see Experimental Setup). As
shown in Table~\ref{tab:benchmark}, RVT$^*$ has weaker initial performance than the vanilla RVT
model, mainly due to the Performer approximation of the self-attention.%

In Figure~\ref{fig:att_robust}, we visualize the attention maps of self-attention and \texttt{\model-S,-D,-SD} under different image corruption scenarios. We see
that for a clean image, all the attention designs can roughly locate salient regions around the main object (airplane). However, self-attention only highlights the center part of the object while \texttt{\model-SD} and \texttt{\model-D} capture the contour of the object more clearly.
For corrupted images, we observe that attention maps from vanilla self-attention tend to be noisier than \model. For example, with severe
weather corruption (the last row), self-attention misses the main object and rather highlights the snow effect, while \texttt{\model-SD} still captures
the object and suppresses the noise. We also notice that \texttt{\model-S} tend to produce a blurrier attention map compared to the ones with a dynamic dictionary, which might be due to the weaker expressivity of static dictionary compared to the dynamic dictionary.

\begin{table}[t]
\caption{\textbf{Evaluation of domain generalization on PACS.} Our \texttt{\model-SD} outperforms the baseline RVT$^*$ and other variants.\vspace{1mm}}
\label{tab:domain_gen}
\centering
\begin{small}
\adjustbox{width=\linewidth}{
\begin{tabular}{lP{0.08\textwidth}P{0.08\textwidth}P{0.08\textwidth}P{0.08\textwidth}}
\toprule
Target & Photo & Sketch & Cartoon & Art \\
\midrule\midrule
RVT$^\ast$ & 94.19 & 81.73 & 79.78 & 81.25 \\
\model-S & 93.89 & 82.62 & 80.16 & 81.49 \\
\model-D & 96.29 & 80.40 & 80.33 & 84.77 \\
\model-SD & \textbf{96.47} & \textbf{82.78} & \textbf{80.98} & \textbf{86.08} \\
\bottomrule
\end{tabular}}
\end{small}
\vskip -0.2in
\end{table}

\subsection{Evaluation on Domain Generalization}
\label{sec:exp_dg}
Domain generalization is a related setting, which evaluates the models' generalization to unseen domains at test time.
Here we finetune the ImageNet-pretrained models on three source domains in PACS~\cite{li2017deeper} and test them on the left-out target domain.

Table~\ref{tab:domain_gen} shows that \texttt{\model-SD} outperforms the RVT baseline across all four target domains. Specifically, \texttt{\model-SD} improves RVT$^*$ from 81.25\% to 86.08\% on the \emph{Art} domain and from 94.19\% to 96.47\% on the \emph{Photo} domain. These results indicate that our attention module is more robust than self-attention
when generalizing to unseen domains.

\begin{table}[t]
\caption{\textbf{Segmentation evaluation on PASCAL VOC using attention maps}. Our \texttt{\model-SD} improves the mean IOU score of the baseline RVT$^*$ and is more selective (higher FN scores). \vspace{1mm} }
\label{tab:segmentation}
\centering
\begin{small}
\adjustbox{width=\linewidth}{
\begin{tabular}{lP{0.09\textwidth}P{0.09\textwidth}P{0.09\textwidth}P{0.09\textwidth}}
\toprule
 & RVT* & \model-S & \model-D & \model-SD \\
\midrule\midrule
mIoU$\uparrow$ & 39.92 & 43.33 & 42.03 & \textbf{44.15} \\
FP$\downarrow$ & 49.41 & \textbf{23.11}& 25.23 & 29.28 \\
FN$\downarrow$ & \textbf{3.95} & 12.08 & 11.77 & {8.76} \\
\bottomrule
\end{tabular}}
\end{small}
\vskip -0.2in
\end{table}

\subsection{Evaluation on Segmentation and Eye Fixation}
\label{sec:exp_seg_eye}

\minisection{Attention as coarse image segmentation}. Recently, self-supervised vision transformers~\cite{caron2021emerging} have been shown to produce attention maps that are similar to the semantic segmentation of foreground objects. Following ~\citet{caron2021emerging}, we evaluate RVT$^*$ with self-attention and \model on the validation set of PASCAL VOC 2012 using the model trained on ImageNet-1K. To obtain a segmentation map, we normalize an attention map from the global average of tokens to [0, 1] and use a threshold 0.3 to distinguish foreground objects from the background (class agnostic). The main evaluation metric is mean IoU which evaluates the overlapping area between a predicted segmentation map and the ground truth. We also consider false positive (FP) and false negative (FN) rates as metrics.

Table~\ref{tab:segmentation} shows that all three variants of \model achieve higher mean IoU compared to the self-attention counterpart RVT$^*$, where \texttt{\model-SD} improves the score from 39.92\% to
44.15\%. Also, the FP rate is substantially reduced by our attention framework, indicating that \model can effectively filter out distracting
information and preserve only the relevant information about the foreground objects. Another observation is that \model has a higher FN rate, suggesting \model is more selective than self-attention and emphasize more on the core parts of the objects.

\begin{table}[t]
\caption{\textbf{Evaluation on human eye fixations.} Here our \texttt{\model-S} achieves the highest score while all variants outperforms RVT$^*$.\vspace{1mm}}
\label{table:eye_fixation}
\centering
\begin{small}
\adjustbox{width=\linewidth}{
\begin{tabular}{cP{0.09\textwidth}P{0.09\textwidth}P{0.09\textwidth}P{0.09\textwidth}}
\toprule
 Metric & RVT$^*$ & \model-S & \model-D & \model-SD \\
\midrule\midrule
NSS & 0.502 & \textbf{0.737} & 0.632 & 0.678 \\
\bottomrule
\end{tabular}}
\end{small}
\end{table}

\begin{figure}[t]
\begin{center}
\centerline{\includegraphics[width=0.9\columnwidth]{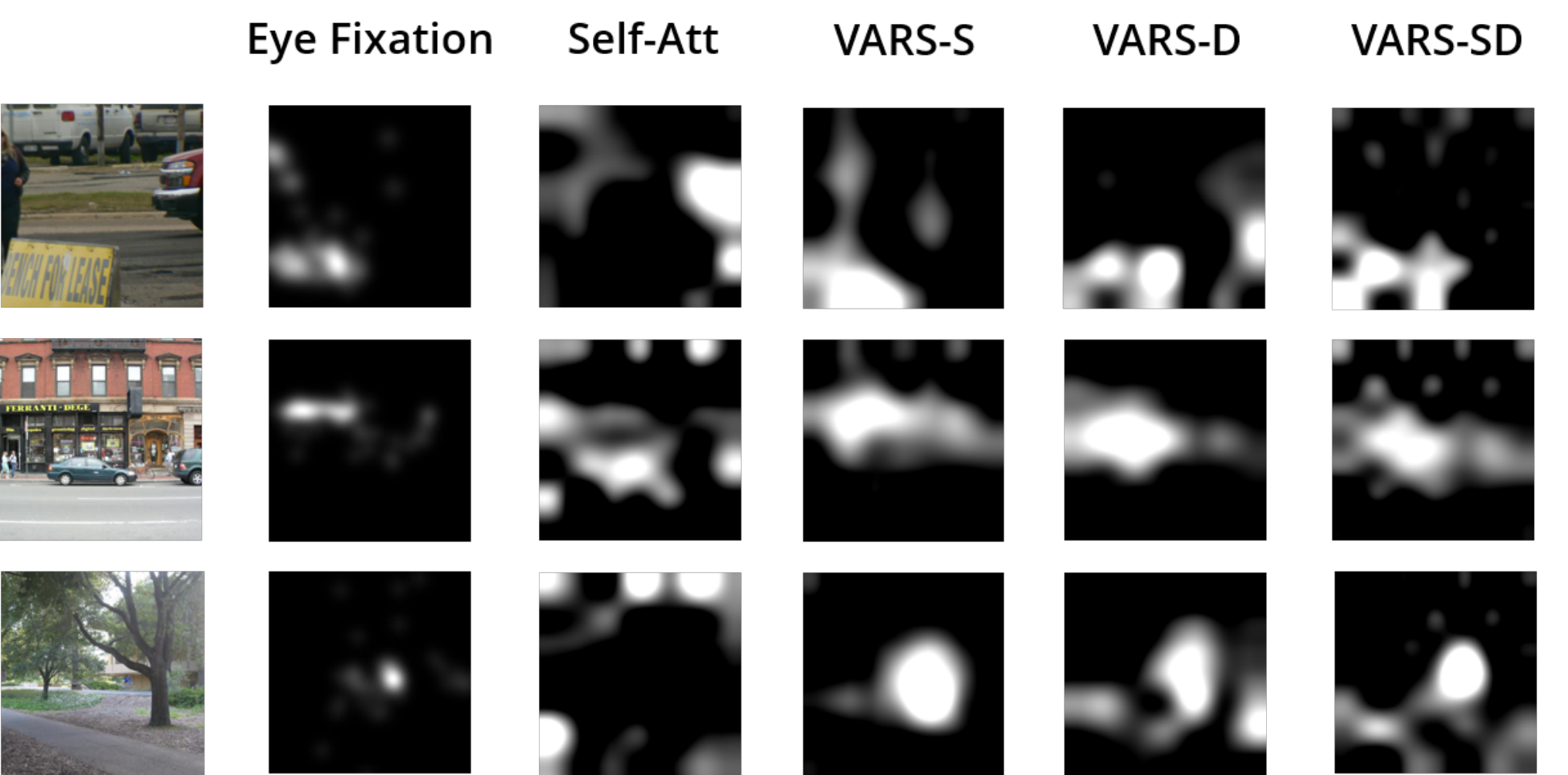}}
\vskip -0.1in
\caption{\textbf{Visualization on eye fixation.} \model' attention maps are more consistent with human eye fixation than self-attention's. }
\label{fig:att_eye_fix}
\end{center}
\vskip -0.3in
\end{figure}

\textbf{Alignment with human eye fixations.} Since human eye fixation is under the guidance of bottom-up attention~\cite{zhaoping2014understanding}, here we investigate how close our attention maps are to the human eye fixation maps. Here we evaluate the ImageNet-pretrained RVT with self-attention and \model on MIT1003~\cite{judd2009learning}, containing 1K natural images with eye fixation maps collected from 15 human observers. We adopt the metric of normalized scanpath saliency (NSS)~\cite{peters2005components} that measures an average of normalized attention value at fixated positions.

Table~\ref{table:eye_fixation} shows that RVT with \model achieves higher NSS scores than RVT with self-attention (\ie RVT$^*$),  aligning better with the human eye fixations data. Figure~\ref{fig:att_eye_fix} shows the attention maps captured from humans and generated by the models. We notice that \model predicts regions that are more closely aligned with human attention, while self-attention tend to highlight irrelevant background regions.

\subsection{Analysis and Ablation Study}
\label{sec:ablation}
\begin{figure}[t]
\centering
\centerline{\includegraphics[width=0.8\columnwidth]{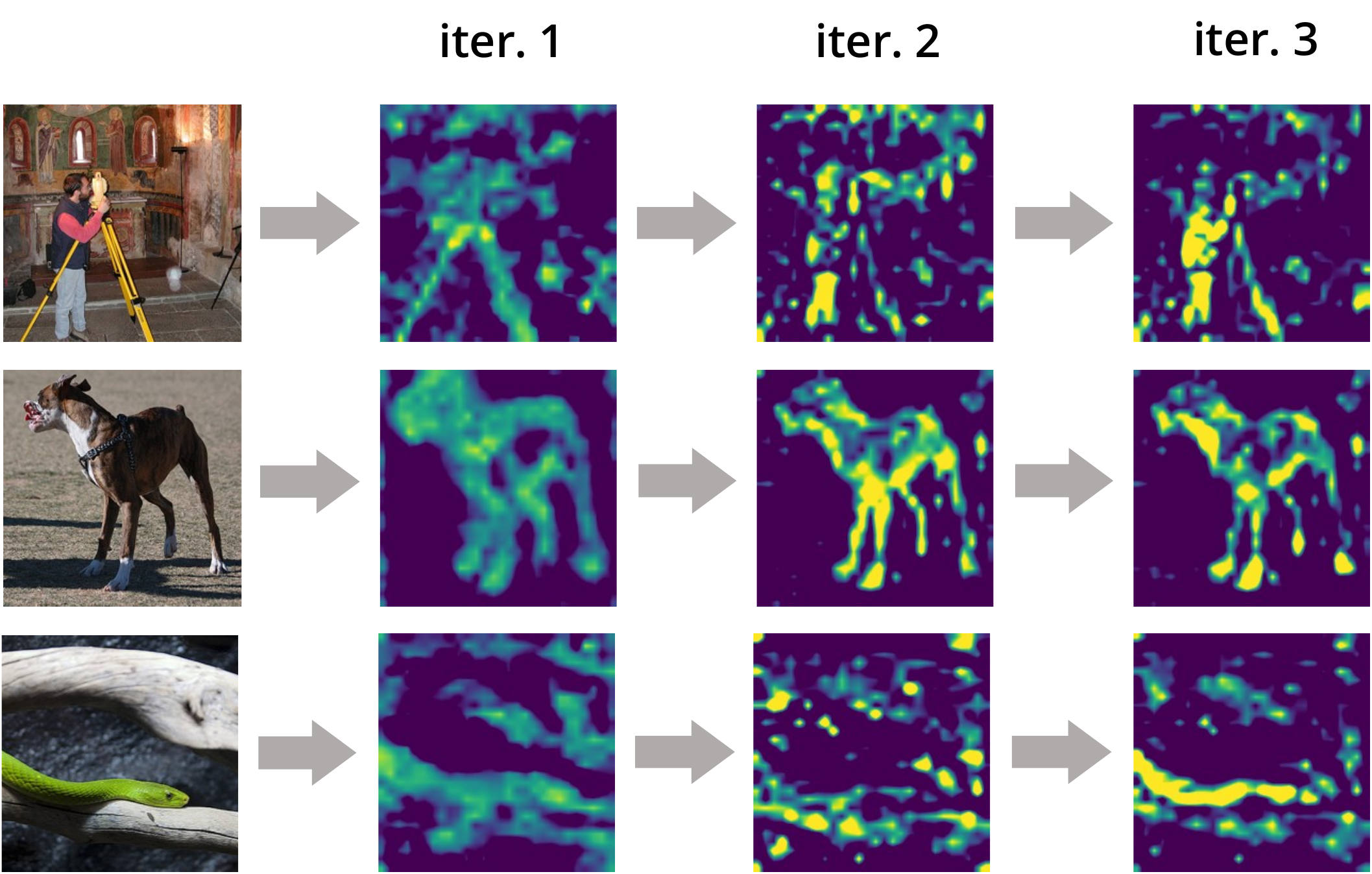}}
\vskip -0.1in
\caption{\textbf{Recurrent refinement of attention maps.} \model refines the attention maps iteratively during the recurrent updates.}
\label{fig:att_each_iter}
\end{figure}

\minisection{Recurrent refinement of attention.} \model performs recurrent sparse reconstruction of the inputs in an iterative manner. In Figure~\ref{fig:att_each_iter}, we visualize the attention maps \texttt{\model-S} on ImageNet validation samples at different updating steps. %
We can see that \model refines the attention maps through recurrent updates, i.e., the attention maps become more focused on the core parts of the objects while suppressing the background and other distracting objects.

\begin{figure}[t]
\centering
\centerline{\includegraphics[width=0.9\columnwidth]{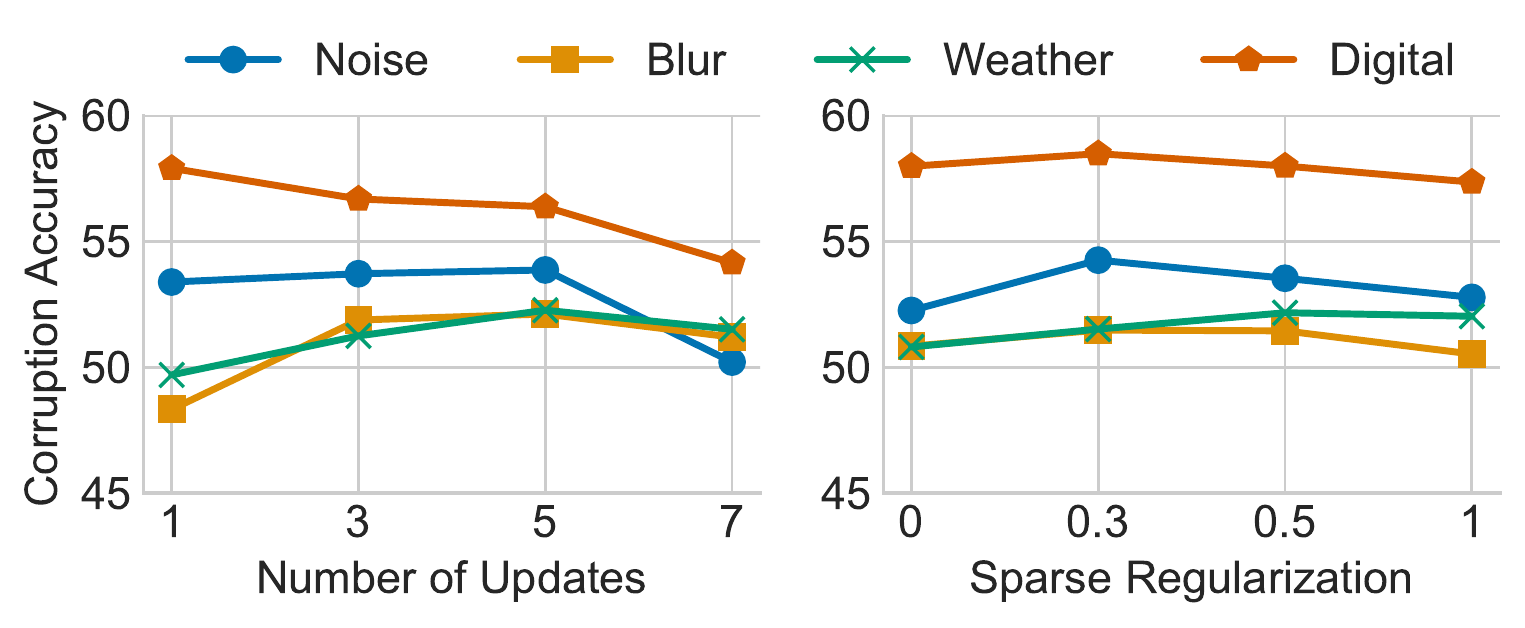}}
\vskip -0.1in
\caption{\textbf{Hyperparameter analysis.} We study the number of updates (left) and the level of sparse regularization (right) in sparse reconstruction. \model performs similarly with 3 to 5 iteration steps and we choose 3 for better efficiency. \model is not sensitive to the level of sparse regularization and we use 0.3 in the experiments. }
\label{fig:ablation_iter_sparse}
\vskip -0.2in
\end{figure}

\minisection{Number of recurrent updates.} Figure~\ref{fig:ablation_iter_sparse} (left) shows the accuracy on ImageNet-C over different number of updates $k$. We find that the model has a similar performance between $k=3$ and $5$ with a drop of performance at $k=1$ and $7$. We choose $k=3$ in our experiments for efficiency.

\minisection{Strength of sparse constraints.} Figure~\ref{fig:ablation_iter_sparse} (right) shows the accuracy over different $\lambda$ values that determine the level of sparse regularization during the reconstruction of input. We observe that the curves are relatively flat which indicates \model is not very sensitive to the strength of the sparse regularization. We adopt $\lambda=0.3$ in our experiments which has a slightly better performance than the other values.

\section{Conclusion}
We introduced a new attention formulation---Visual Attention from Recurrent Sparse reconstruction (\model)---which takes inspiration from the robustness 
characteristics of human vision. We
observed a connection among visual attention, recurrency, and sparsity and showed that contemporary attention models can be derived from recurrent sparse reconstruction of input
signals. \model adopts an ODE based formulation to describe  neural dynamics; equilibrium states are solved by iteratively optimizing the sparse reconstruction of input. We showed that self-attention is a special case of \model with approximate neural dynamics and no sparsity constraints. \model is a general attention module that can be plugged into vision transformers, replacing the self-attention module, offering improved performance. We conducted extensive evaluation on five robustness benchmarks and three additional datasets of related settings to understand the properties of \model. We found \model increases model robustness with improved quality of attention maps across various datasets and settings.

\bibliography{example_paper}

\begin{thebibliography}{91}
\providecommand{\natexlab}[1]{#1}
\providecommand{\url}[1]{\texttt{#1}}
\expandafter\ifx\csname urlstyle\endcsname\relax
  \providecommand{\doi}[1]{doi: #1}\else
  \providecommand{\doi}{doi: \begingroup \urlstyle{rm}\Url}\fi

\bibitem[Bai et~al.(2019)Bai, Kolter, and Koltun]{bai2019deep}
Bai, S., Kolter, J.~Z., and Koltun, V.
\newblock Deep equilibrium models.
\newblock \emph{arXiv preprint arXiv:1909.01377}, 2019.

\bibitem[Beck \& Teboulle(2009)Beck and Teboulle]{beck2009fast}
Beck, A. and Teboulle, M.
\newblock A fast iterative shrinkage-thresholding algorithm for linear inverse
  problems.
\newblock \emph{SIAM journal on imaging sciences}, 2\penalty0 (1):\penalty0
  183--202, 2009.

\bibitem[Biederman(1987)]{biederman1987recognition}
Biederman, I.
\newblock Recognition-by-components: a theory of human image understanding.
\newblock \emph{Psychological review}, 94\penalty0 (2):\penalty0 115, 1987.

\bibitem[Bisanz et~al.(2012)Bisanz, Bisanz, and Kail]{bisanz2012learning}
Bisanz, J., Bisanz, G.~L., and Kail, R.
\newblock \emph{Learning in children: Progress in cognitive development
  research}.
\newblock Springer Science \& Business Media, 2012.

\bibitem[Bosking et~al.(1997)Bosking, Zhang, Schofield, and
  Fitzpatrick]{bosking1997orientation}
Bosking, W.~H., Zhang, Y., Schofield, B., and Fitzpatrick, D.
\newblock Orientation selectivity and the arrangement of horizontal connections
  in tree shrew striate cortex.
\newblock \emph{Journal of neuroscience}, 17\penalty0 (6):\penalty0 2112--2127,
  1997.

\bibitem[Caron et~al.(2021)Caron, Touvron, Misra, J{\'e}gou, Mairal,
  Bojanowski, and Joulin]{caron2021emerging}
Caron, M., Touvron, H., Misra, I., J{\'e}gou, H., Mairal, J., Bojanowski, P.,
  and Joulin, A.
\newblock Emerging properties in self-supervised vision transformers.
\newblock \emph{arXiv preprint arXiv:2104.14294}, 2021.

\bibitem[Chaney et~al.(2014)Chaney, Fischer, and
  Whitney]{chaney2014hierarchical}
Chaney, W., Fischer, J., and Whitney, D.
\newblock The hierarchical sparse selection model of visual crowding.
\newblock \emph{Frontiers in integrative neuroscience}, 8:\penalty0 73, 2014.

\bibitem[Choromanski et~al.(2020)Choromanski, Likhosherstov, Dohan, Song, Gane,
  Sarlos, Hawkins, Davis, Mohiuddin, Kaiser, et~al.]{choromanski2020rethinking}
Choromanski, K., Likhosherstov, V., Dohan, D., Song, X., Gane, A., Sarlos, T.,
  Hawkins, P., Davis, J., Mohiuddin, A., Kaiser, L., et~al.
\newblock Rethinking attention with performers.
\newblock \emph{arXiv preprint arXiv:2009.14794}, 2020.

\bibitem[Cohen \& Grossberg(1983)Cohen and Grossberg]{cohen1983absolute}
Cohen, M.~A. and Grossberg, S.
\newblock Absolute stability of global pattern formation and parallel memory
  storage by competitive neural networks.
\newblock \emph{IEEE transactions on systems, man, and cybernetics}, \penalty0
  (5):\penalty0 815--826, 1983.

\bibitem[Darrell et~al.(1990)Darrell, Sclaroff, and
  Pentland]{darrell1990segmentation}
Darrell, T., Sclaroff, S., and Pentland, A.
\newblock Segmentation by minimal description.
\newblock In \emph{Proceedings Third International Conference on Computer
  Vision}, pp.\  112--113. IEEE Computer Society, 1990.

\bibitem[d'Ascoli et~al.(2021)d'Ascoli, Touvron, Leavitt, Morcos, Biroli, and
  Sagun]{d2021convit}
d'Ascoli, S., Touvron, H., Leavitt, M., Morcos, A., Biroli, G., and Sagun, L.
\newblock Convit: Improving vision transformers with soft convolutional
  inductive biases.
\newblock \emph{arXiv preprint arXiv:2103.10697}, 2021.

\bibitem[Dayan \& Abbott(2001)Dayan and Abbott]{dayan2001theoretical}
Dayan, P. and Abbott, L.~F.
\newblock \emph{Theoretical neuroscience: computational and mathematical
  modeling of neural systems}.
\newblock Computational Neuroscience Series, 2001.

\bibitem[Deng et~al.(2009)Deng, Dong, Socher, Li, Li, and
  Fei-Fei]{deng2009imagenet}
Deng, J., Dong, W., Socher, R., Li, L.-J., Li, K., and Fei-Fei, L.
\newblock Imagenet: A large-scale hierarchical image database.
\newblock In \emph{IEEE conference on computer vision and pattern recognition},
  2009.

\bibitem[Desimone \& Duncan(1995)Desimone and Duncan]{desimone1995neural}
Desimone, R. and Duncan, J.
\newblock Neural mechanisms of selective visual attention.
\newblock \emph{Annual review of neuroscience}, 18\penalty0 (1):\penalty0
  193--222, 1995.

\bibitem[Djolonga et~al.(2021)Djolonga, Yung, Tschannen, Romijnders, Beyer,
  Kolesnikov, Puigcerver, Minderer, D'Amour, Moldovan,
  et~al.]{djolonga2021robustness}
Djolonga, J., Yung, J., Tschannen, M., Romijnders, R., Beyer, L., Kolesnikov,
  A., Puigcerver, J., Minderer, M., D'Amour, A., Moldovan, D., et~al.
\newblock On robustness and transferability of convolutional neural networks.
\newblock In \emph{Proceedings of the IEEE/CVF Conference on Computer Vision
  and Pattern Recognition}, pp.\  16458--16468, 2021.

\bibitem[Dodge \& Karam(2017)Dodge and Karam]{dodge2017study}
Dodge, S. and Karam, L.
\newblock A study and comparison of human and deep learning recognition
  performance under visual distortions.
\newblock In \emph{2017 26th international conference on computer communication
  and networks (ICCCN)}, pp.\  1--7. IEEE, 2017.

\bibitem[Dong et~al.(2020)Dong, Li, Wang, and Xu]{dong2020adversarially}
Dong, M., Li, Y., Wang, Y., and Xu, C.
\newblock Adversarially robust neural architectures.
\newblock \emph{arXiv preprint arXiv:2009.00902}, 2020.

\bibitem[Dosovitskiy et~al.(2020)Dosovitskiy, Beyer, Kolesnikov, Weissenborn,
  Zhai, Unterthiner, Dehghani, Minderer, Heigold, Gelly,
  et~al.]{dosovitskiy2020image}
Dosovitskiy, A., Beyer, L., Kolesnikov, A., Weissenborn, D., Zhai, X.,
  Unterthiner, T., Dehghani, M., Minderer, M., Heigold, G., Gelly, S., et~al.
\newblock An image is worth 16x16 words: Transformers for image recognition at
  scale.
\newblock \emph{arXiv preprint arXiv:2010.11929}, 2020.

\bibitem[Everingham et~al.(2010)Everingham, Van~Gool, Williams, Winn, and
  Zisserman]{everingham2010pascal}
Everingham, M., Van~Gool, L., Williams, C.~K., Winn, J., and Zisserman, A.
\newblock The pascal visual object classes (voc) challenge.
\newblock \emph{IJCV}, 88\penalty0 (2):\penalty0 303--338, 2010.

\bibitem[Fan et~al.(2020)Fan, Yu, Mei, Zhang, Fu, Liu, and
  Huang]{fan2020neural}
Fan, Y., Yu, J., Mei, Y., Zhang, Y., Fu, Y., Liu, D., and Huang, T.~S.
\newblock Neural sparse representation for image restoration.
\newblock \emph{arXiv preprint arXiv:2006.04357}, 2020.

\bibitem[Felleman \& Van~Essen(1991)Felleman and
  Van~Essen]{felleman1991distributed}
Felleman, D.~J. and Van~Essen, D.~C.
\newblock Distributed hierarchical processing in the primate cerebral cortex.
\newblock \emph{Cerebral cortex (New York, NY: 1991)}, 1\penalty0 (1):\penalty0
  1--47, 1991.

\bibitem[Geirhos et~al.(2017)Geirhos, Janssen, Sch{\"u}tt, Rauber, Bethge, and
  Wichmann]{geirhos2017comparing}
Geirhos, R., Janssen, D.~H., Sch{\"u}tt, H.~H., Rauber, J., Bethge, M., and
  Wichmann, F.~A.
\newblock Comparing deep neural networks against humans: object recognition
  when the signal gets weaker.
\newblock \emph{arXiv preprint arXiv:1706.06969}, 2017.

\bibitem[Geirhos et~al.(2018)Geirhos, Rubisch, Michaelis, Bethge, Wichmann, and
  Brendel]{geirhos2018imagenet}
Geirhos, R., Rubisch, P., Michaelis, C., Bethge, M., Wichmann, F.~A., and
  Brendel, W.
\newblock Imagenet-trained cnns are biased towards texture; increasing shape
  bias improves accuracy and robustness.
\newblock \emph{arXiv preprint arXiv:1811.12231}, 2018.

\bibitem[Gilbert \& Wiesel(1989)Gilbert and Wiesel]{gilbert1989columnar}
Gilbert, C.~D. and Wiesel, T.~N.
\newblock Columnar specificity of intrinsic horizontal and corticocortical
  connections in cat visual cortex.
\newblock \emph{Journal of Neuroscience}, 9\penalty0 (7):\penalty0 2432--2442,
  1989.

\bibitem[Grossberg \& Mingolla(1987)Grossberg and
  Mingolla]{grossberg1987neural}
Grossberg, S. and Mingolla, E.
\newblock Neural dynamics of perceptual grouping: Textures, boundaries, and
  emergent segmentations.
\newblock In \emph{The adaptive brain II}, pp.\  143--210. Elsevier, 1987.

\bibitem[Guo et~al.(2018)Guo, Zhang, Zhang, and Chen]{guo2018sparse}
Guo, Y., Zhang, C., Zhang, C., and Chen, Y.
\newblock Sparse dnns with improved adversarial robustness.
\newblock \emph{arXiv preprint arXiv:1810.09619}, 2018.

\bibitem[He et~al.(2015)He, Zhang, Ren, and Sun]{he2015delving}
He, K., Zhang, X., Ren, S., and Sun, J.
\newblock Delving deep into rectifiers: Surpassing human-level performance on
  imagenet classification.
\newblock In \emph{Proceedings of the IEEE international conference on computer
  vision}, pp.\  1026--1034, 2015.

\bibitem[He et~al.(2016)He, Zhang, Ren, and Sun]{he2016deep}
He, K., Zhang, X., Ren, S., and Sun, J.
\newblock Deep residual learning for image recognition.
\newblock In \emph{Proceedings of the IEEE conference on computer vision and
  pattern recognition}, pp.\  770--778, 2016.

\bibitem[Hendrycks \& Dietterich(2019)Hendrycks and
  Dietterich]{hendrycks2019benchmarking}
Hendrycks, D. and Dietterich, T.
\newblock Benchmarking neural network robustness to common corruptions and
  perturbations.
\newblock \emph{arXiv preprint arXiv:1903.12261}, 2019.

\bibitem[Hendrycks et~al.(2021{\natexlab{a}})Hendrycks, Basart, Mu, Kadavath,
  Wang, Dorundo, Desai, Zhu, Parajuli, Guo, et~al.]{hendrycks2021many}
Hendrycks, D., Basart, S., Mu, N., Kadavath, S., Wang, F., Dorundo, E., Desai,
  R., Zhu, T., Parajuli, S., Guo, M., et~al.
\newblock The many faces of robustness: A critical analysis of
  out-of-distribution generalization.
\newblock In \emph{Proceedings of the IEEE/CVF International Conference on
  Computer Vision}, pp.\  8340--8349, 2021{\natexlab{a}}.

\bibitem[Hendrycks et~al.(2021{\natexlab{b}})Hendrycks, Zhao, Basart,
  Steinhardt, and Song]{hendrycks2021natural}
Hendrycks, D., Zhao, K., Basart, S., Steinhardt, J., and Song, D.
\newblock Natural adversarial examples.
\newblock In \emph{Proceedings of the IEEE/CVF Conference on Computer Vision
  and Pattern Recognition}, pp.\  15262--15271, 2021{\natexlab{b}}.

\bibitem[Heo et~al.(2021)Heo, Yun, Han, Chun, Choe, and Oh]{heo2021rethinking}
Heo, B., Yun, S., Han, D., Chun, S., Choe, J., and Oh, S.~J.
\newblock Rethinking spatial dimensions of vision transformers.
\newblock \emph{arXiv preprint arXiv:2103.16302}, 2021.

\bibitem[Hochreiter \& Schmidhuber(1997)Hochreiter and
  Schmidhuber]{hochreiter1997long}
Hochreiter, S. and Schmidhuber, J.
\newblock Long short-term memory.
\newblock \emph{Neural computation}, 9\penalty0 (8):\penalty0 1735--1780, 1997.

\bibitem[Hoefler et~al.(2021)Hoefler, Alistarh, Ben-Nun, Dryden, and
  Peste]{hoefler2021sparsity}
Hoefler, T., Alistarh, D., Ben-Nun, T., Dryden, N., and Peste, A.
\newblock Sparsity in deep learning: Pruning and growth for efficient inference
  and training in neural networks.
\newblock \emph{arXiv preprint arXiv:2102.00554}, 2021.

\bibitem[Hopfield(1984)]{hopfield1984neurons}
Hopfield, J.~J.
\newblock Neurons with graded response have collective computational properties
  like those of two-state neurons.
\newblock \emph{Proceedings of the national academy of sciences}, 81\penalty0
  (10):\penalty0 3088--3092, 1984.

\bibitem[Huang et~al.(2020)Huang, Gornet, Dai, Yu, Nguyen, Tsao, and
  Anandkumar]{huang2020neural}
Huang, Y., Gornet, J., Dai, S., Yu, Z., Nguyen, T., Tsao, D.~Y., and
  Anandkumar, A.
\newblock Neural networks with recurrent generative feedback.
\newblock \emph{arXiv preprint arXiv:2007.09200}, 2020.

\bibitem[Itti \& Koch(2001)Itti and Koch]{itti2001computational}
Itti, L. and Koch, C.
\newblock Computational modelling of visual attention.
\newblock \emph{Nature reviews neuroscience}, 2\penalty0 (3):\penalty0
  194--203, 2001.

\bibitem[Judd et~al.(2009)Judd, Ehinger, Durand, and
  Torralba]{judd2009learning}
Judd, T., Ehinger, K., Durand, F., and Torralba, A.
\newblock Learning to predict where humans look.
\newblock In \emph{2009 IEEE 12th international conference on computer vision},
  pp.\  2106--2113. IEEE, 2009.

\bibitem[Kar et~al.(2019)Kar, Kubilius, Schmidt, Issa, and
  DiCarlo]{kar2019evidence}
Kar, K., Kubilius, J., Schmidt, K., Issa, E.~B., and DiCarlo, J.~J.
\newblock Evidence that recurrent circuits are critical to the ventral
  stream’s execution of core object recognition behavior.
\newblock \emph{Nature neuroscience}, 22\penalty0 (6):\penalty0 974--983, 2019.

\bibitem[Kim et~al.(2019)Kim, Linsley, Thakkar, and
  Serre]{kim2019disentangling}
Kim, J., Linsley, D., Thakkar, K., and Serre, T.
\newblock Disentangling neural mechanisms for perceptual grouping.
\newblock \emph{arXiv preprint arXiv:1906.01558}, 2019.

\bibitem[Koch \& Ullman(1987)Koch and Ullman]{koch1987shifts}
Koch, C. and Ullman, S.
\newblock Shifts in selective visual attention: towards the underlying neural
  circuitry.
\newblock In \emph{Matters of intelligence}, pp.\  115--141. Springer, 1987.

\bibitem[Lamme \& Roelfsema(2000)Lamme and Roelfsema]{lamme2000distinct}
Lamme, V.~A. and Roelfsema, P.~R.
\newblock The distinct modes of vision offered by feedforward and recurrent
  processing.
\newblock \emph{Trends in neurosciences}, 23\penalty0 (11):\penalty0 571--579,
  2000.

\bibitem[Lee et~al.(2007)Lee, Ekanadham, and Ng]{lee2007sparse}
Lee, H., Ekanadham, C., and Ng, A.
\newblock Sparse deep belief net model for visual area v2.
\newblock \emph{Advances in neural information processing systems},
  20:\penalty0 873--880, 2007.

\bibitem[Li et~al.(2017)Li, Yang, Song, and Hospedales]{li2017deeper}
Li, D., Yang, Y., Song, Y.-Z., and Hospedales, T.~M.
\newblock Deeper, broader and artier domain generalization.
\newblock In \emph{Proceedings of the IEEE international conference on computer
  vision}, pp.\  5542--5550, 2017.

\bibitem[Li(1998)]{li1998neural}
Li, Z.
\newblock A neural model of contour integration in the primary visual cortex.
\newblock \emph{Neural computation}, 10\penalty0 (4):\penalty0 903--940, 1998.

\bibitem[Li(2014)]{zhaoping2014understanding}
Li, Z.
\newblock \emph{Understanding vision: theory, models, and data}.
\newblock Oxford University Press, USA, 2014.

\bibitem[Linsley et~al.(2020)Linsley, Kim, Ashok, and
  Serre]{linsley2020recurrent}
Linsley, D., Kim, J., Ashok, A., and Serre, T.
\newblock Recurrent neural circuits for contour detection.
\newblock \emph{arXiv preprint arXiv:2010.15314}, 2020.

\bibitem[Liu et~al.(2019)Liu, Mocanu, and Pechenizkiy]{liu2019improving}
Liu, S., Mocanu, D.~C., and Pechenizkiy, M.
\newblock On improving deep learning generalization with adaptive sparse
  connectivity.
\newblock \emph{arXiv preprint arXiv:1906.11626}, 2019.

\bibitem[Machiraju et~al.(2021)Machiraju, Choung, Frossard, Herzog,
  et~al.]{machiraju2021bio}
Machiraju, H., Choung, O.-H., Frossard, P., Herzog, M., et~al.
\newblock Bio-inspired robustness: A review.
\newblock \emph{arXiv preprint arXiv:2103.09265}, 2021.

\bibitem[Madry et~al.(2017)Madry, Makelov, Schmidt, Tsipras, and
  Vladu]{madry2017towards}
Madry, A., Makelov, A., Schmidt, L., Tsipras, D., and Vladu, A.
\newblock Towards deep learning models resistant to adversarial attacks.
\newblock \emph{arXiv preprint arXiv:1706.06083}, 2017.

\bibitem[Mao et~al.(2021)Mao, Qi, Chen, Li, Duan, Ye, He, and
  Xue]{mao2021towards}
Mao, X., Qi, G., Chen, Y., Li, X., Duan, R., Ye, S., He, Y., and Xue, H.
\newblock Towards robust vision transformer.
\newblock \emph{arXiv preprint arXiv:2105.07926}, 2021.

\bibitem[M{\'e}ly et~al.(2018)M{\'e}ly, Linsley, and
  Serre]{mely2018complementary}
M{\'e}ly, D.~A., Linsley, D., and Serre, T.
\newblock Complementary surrounds explain diverse contextual phenomena across
  visual modalities.
\newblock \emph{Psychological review}, 125\penalty0 (5):\penalty0 769, 2018.

\bibitem[Mnih et~al.(2014)Mnih, Heess, Graves, et~al.]{mnih2014recurrent}
Mnih, V., Heess, N., Graves, A., et~al.
\newblock Recurrent models of visual attention.
\newblock In \emph{Advances in neural information processing systems}, pp.\
  2204--2212, 2014.

\bibitem[Naseer et~al.(2021)Naseer, Ranasinghe, Khan, Hayat, Khan, and
  Yang]{naseer2021intriguing}
Naseer, M., Ranasinghe, K., Khan, S., Hayat, M., Khan, F.~S., and Yang, M.-H.
\newblock Intriguing properties of vision transformers.
\newblock \emph{arXiv preprint arXiv:2105.10497}, 2021.

\bibitem[Nayebi et~al.(2018)Nayebi, Bear, Kubilius, Kar, Ganguli, Sussillo,
  DiCarlo, and Yamins]{nayebi2018task}
Nayebi, A., Bear, D., Kubilius, J., Kar, K., Ganguli, S., Sussillo, D.,
  DiCarlo, J.~J., and Yamins, D.~L.
\newblock Task-driven convolutional recurrent models of the visual system.
\newblock \emph{arXiv preprint arXiv:1807.00053}, 2018.

\bibitem[Olshausen \& Field(1996)Olshausen and Field]{olshausen1996emergence}
Olshausen, B.~A. and Field, D.~J.
\newblock Emergence of simple-cell receptive field properties by learning a
  sparse code for natural images.
\newblock \emph{Nature}, 381\penalty0 (6583):\penalty0 607--609, 1996.

\bibitem[Olshausen \& Field(1997)Olshausen and Field]{olshausen1997sparse}
Olshausen, B.~A. and Field, D.~J.
\newblock Sparse coding with an overcomplete basis set: A strategy employed by
  v1?
\newblock \emph{Vision research}, 37\penalty0 (23):\penalty0 3311--3325, 1997.

\bibitem[O'Reilly et~al.(2013)O'Reilly, Wyatte, Herd, Mingus, and
  Jilk]{o2013recurrent}
O'Reilly, R.~C., Wyatte, D., Herd, S., Mingus, B., and Jilk, D.~J.
\newblock Recurrent processing during object recognition.
\newblock \emph{Frontiers in psychology}, 4:\penalty0 124, 2013.

\bibitem[Paul \& Chen(2021)Paul and Chen]{paul2021vision}
Paul, S. and Chen, P.-Y.
\newblock Vision transformers are robust learners.
\newblock \emph{arXiv preprint arXiv:2105.07581}, 2021.

\bibitem[Peters et~al.(2005)Peters, Iyer, Itti, and Koch]{peters2005components}
Peters, R.~J., Iyer, A., Itti, L., and Koch, C.
\newblock Components of bottom-up gaze allocation in natural images.
\newblock \emph{Vision research}, 45\penalty0 (18):\penalty0 2397--2416, 2005.

\bibitem[Radosavovic et~al.(2020)Radosavovic, Kosaraju, Girshick, He, and
  Doll{\'a}r]{radosavovic2020designing}
Radosavovic, I., Kosaraju, R.~P., Girshick, R., He, K., and Doll{\'a}r, P.
\newblock Designing network design spaces.
\newblock In \emph{Proceedings of the IEEE/CVF Conference on Computer Vision
  and Pattern Recognition}, pp.\  10428--10436, 2020.

\bibitem[Rao et~al.(2021)Rao, Zhao, Zhu, Lu, and Zhou]{rao2021global}
Rao, Y., Zhao, W., Zhu, Z., Lu, J., and Zhou, J.
\newblock Global filter networks for image classification.
\newblock \emph{arXiv preprint arXiv:2107.00645}, 2021.

\bibitem[Rebuffi et~al.(2021)Rebuffi, Gowal, Calian, Stimberg, Wiles, and
  Mann]{rebuffi2021data}
Rebuffi, S.-A., Gowal, S., Calian, D.~A., Stimberg, F., Wiles, O., and Mann,
  T.~A.
\newblock Data augmentation can improve robustness.
\newblock \emph{Advances in Neural Information Processing Systems}, 34, 2021.

\bibitem[Roelfsema(2006)]{roelfsema2006cortical}
Roelfsema, P.~R.
\newblock Cortical algorithms for perceptual grouping.
\newblock \emph{Annu. Rev. Neurosci.}, 29:\penalty0 203--227, 2006.

\bibitem[Roelfsema et~al.(2002)Roelfsema, Lamme, Spekreijse, and
  Bosch]{roelfsema2002figure}
Roelfsema, P.~R., Lamme, V.~A., Spekreijse, H., and Bosch, H.
\newblock Figure—ground segregation in a recurrent network architecture.
\newblock \emph{Journal of cognitive neuroscience}, 14\penalty0 (4):\penalty0
  525--537, 2002.

\bibitem[Rozell et~al.(2008)Rozell, Johnson, Baraniuk, and
  Olshausen]{rozell2008sparse}
Rozell, C.~J., Johnson, D.~H., Baraniuk, R.~G., and Olshausen, B.~A.
\newblock Sparse coding via thresholding and local competition in neural
  circuits.
\newblock \emph{Neural computation}, 20\penalty0 (10):\penalty0 2526--2563,
  2008.

\bibitem[Scholl(2001)]{scholl2001objects}
Scholl, B.~J.
\newblock Objects and attention: The state of the art.
\newblock \emph{Cognition}, 80\penalty0 (1-2):\penalty0 1--46, 2001.

\bibitem[Shi et~al.(2020)Shi, Zhang, Dai, Zhu, Mu, and
  Wang]{shi2020informative}
Shi, B., Zhang, D., Dai, Q., Zhu, Z., Mu, Y., and Wang, J.
\newblock Informative dropout for robust representation learning: A shape-bias
  perspective.
\newblock In \emph{International Conference on Machine Learning}, pp.\
  8828--8839. PMLR, 2020.

\bibitem[Sood et~al.(2020)Sood, Tannert, Frassinelli, Bulling, and
  Vu]{sood2020interpreting}
Sood, E., Tannert, S., Frassinelli, D., Bulling, A., and Vu, N.~T.
\newblock Interpreting attention models with human visual attention in machine
  reading comprehension.
\newblock \emph{arXiv preprint arXiv:2010.06396}, 2020.

\bibitem[Stettler et~al.(2002)Stettler, Das, Bennett, and
  Gilbert]{stettler2002lateral}
Stettler, D.~D., Das, A., Bennett, J., and Gilbert, C.~D.
\newblock Lateral connectivity and contextual interactions in macaque primary
  visual cortex.
\newblock \emph{Neuron}, 36\penalty0 (4):\penalty0 739--750, 2002.

\bibitem[Szegedy et~al.(2013)Szegedy, Zaremba, Sutskever, Bruna, Erhan,
  Goodfellow, and Fergus]{szegedy2013intriguing}
Szegedy, C., Zaremba, W., Sutskever, I., Bruna, J., Erhan, D., Goodfellow, I.,
  and Fergus, R.
\newblock Intriguing properties of neural networks.
\newblock \emph{arXiv preprint arXiv:1312.6199}, 2013.

\bibitem[Szegedy et~al.(2016)Szegedy, Vanhoucke, Ioffe, Shlens, and
  Wojna]{szegedy2016rethinking}
Szegedy, C., Vanhoucke, V., Ioffe, S., Shlens, J., and Wojna, Z.
\newblock Rethinking the inception architecture for computer vision.
\newblock In \emph{Proceedings of the IEEE conference on computer vision and
  pattern recognition}, pp.\  2818--2826, 2016.

\bibitem[Tang et~al.(2018)Tang, Schrimpf, Lotter, Moerman, Paredes, Caro,
  Hardesty, Cox, and Kreiman]{tang2018recurrent}
Tang, H., Schrimpf, M., Lotter, W., Moerman, C., Paredes, A., Caro, J.~O.,
  Hardesty, W., Cox, D., and Kreiman, G.
\newblock Recurrent computations for visual pattern completion.
\newblock \emph{Proceedings of the National Academy of Sciences}, 115\penalty0
  (35):\penalty0 8835--8840, 2018.

\bibitem[Touvron et~al.(2021)Touvron, Cord, Douze, Massa, Sablayrolles, and
  J{\'e}gou]{touvron2021training}
Touvron, H., Cord, M., Douze, M., Massa, F., Sablayrolles, A., and J{\'e}gou,
  H.
\newblock Training data-efficient image transformers \& distillation through
  attention.
\newblock In \emph{International Conference on Machine Learning}, pp.\
  10347--10357. PMLR, 2021.

\bibitem[Vaswani et~al.(2017)Vaswani, Shazeer, Parmar, Uszkoreit, Jones, Gomez,
  Kaiser, and Polosukhin]{vaswani2017attention}
Vaswani, A., Shazeer, N., Parmar, N., Uszkoreit, J., Jones, L., Gomez, A.~N.,
  Kaiser, {\L}., and Polosukhin, I.
\newblock Attention is all you need.
\newblock In \emph{Advances in neural information processing systems}, pp.\
  5998--6008, 2017.

\bibitem[Vecera \& Farah(1997)Vecera and Farah]{vecera1997visual}
Vecera, S.~P. and Farah, M.~J.
\newblock Is visual image segmentation a bottom-up or an interactive process?
\newblock \emph{Perception \& Psychophysics}, 59\penalty0 (8):\penalty0
  1280--1296, 1997.

\bibitem[Walther et~al.(2005)Walther, Rutishauser, Koch, and
  Perona]{walther2005selective}
Walther, D., Rutishauser, U., Koch, C., and Perona, P.
\newblock Selective visual attention enables learning and recognition of
  multiple objects in cluttered scenes.
\newblock \emph{Computer Vision and Image Understanding}, 100\penalty0
  (1-2):\penalty0 41--63, 2005.

\bibitem[Wang et~al.(2019)Wang, Ge, Xing, and Lipton]{wang2019learning}
Wang, H., Ge, S., Xing, E.~P., and Lipton, Z.~C.
\newblock Learning robust global representations by penalizing local predictive
  power.
\newblock \emph{arXiv preprint arXiv:1905.13549}, 2019.

\bibitem[Wang et~al.(2021{\natexlab{a}})Wang, Xie, Li, Fan, Song, Liang, Lu,
  Luo, and Shao]{wang2021pyramid}
Wang, W., Xie, E., Li, X., Fan, D.-P., Song, K., Liang, D., Lu, T., Luo, P.,
  and Shao, L.
\newblock Pyramid vision transformer: A versatile backbone for dense prediction
  without convolutions.
\newblock \emph{arXiv preprint arXiv:2102.12122}, 2021{\natexlab{a}}.

\bibitem[Wang et~al.(2018)Wang, Girshick, Gupta, and He]{wang2018non}
Wang, X., Girshick, R., Gupta, A., and He, K.
\newblock Non-local neural networks.
\newblock In \emph{Proceedings of the IEEE conference on computer vision and
  pattern recognition}, pp.\  7794--7803, 2018.

\bibitem[Wang et~al.(2021{\natexlab{b}})Wang, Huang, Liu, Yu, Wang, Gonzalez,
  and Darrell]{wang2021robust}
Wang, X., Huang, T.~E., Liu, B., Yu, F., Wang, X., Gonzalez, J.~E., and
  Darrell, T.
\newblock Robust object detection via instance-level temporal cycle confusion.
\newblock \emph{arXiv preprint arXiv:2104.08381}, 2021{\natexlab{b}}.

\bibitem[Wright et~al.(2008)Wright, Yang, Ganesh, Sastry, and
  Ma]{wright2008robust}
Wright, J., Yang, A.~Y., Ganesh, A., Sastry, S.~S., and Ma, Y.
\newblock Robust face recognition via sparse representation.
\newblock \emph{IEEE transactions on pattern analysis and machine
  intelligence}, 31\penalty0 (2):\penalty0 210--227, 2008.

\bibitem[Wu et~al.(2016)Wu, Wong, Fung, Mi, and Zhang]{wu2016continuous}
Wu, S., Wong, K.~M., Fung, C.~A., Mi, Y., and Zhang, W.
\newblock Continuous attractor neural networks: candidate of a canonical model
  for neural information representation.
\newblock \emph{F1000Research}, 5, 2016.

\bibitem[Wu et~al.(2019)Wu, Rosca, and Lillicrap]{wu2019deep}
Wu, Y., Rosca, M., and Lillicrap, T.
\newblock Deep compressed sensing.
\newblock In \emph{International Conference on Machine Learning}, pp.\
  6850--6860. PMLR, 2019.

\bibitem[Wyatte et~al.(2012)Wyatte, Curran, and O'Reilly]{wyatte2012limits}
Wyatte, D., Curran, T., and O'Reilly, R.
\newblock The limits of feedforward vision: Recurrent processing promotes
  robust object recognition when objects are degraded.
\newblock \emph{Journal of Cognitive Neuroscience}, 24\penalty0 (11):\penalty0
  2248--2261, 2012.

\bibitem[Wyatte et~al.(2014)Wyatte, Jilk, and O'Reilly]{wyatte2014early}
Wyatte, D., Jilk, D.~J., and O'Reilly, R.~C.
\newblock Early recurrent feedback facilitates visual object recognition under
  challenging conditions.
\newblock \emph{Frontiers in psychology}, 5:\penalty0 674, 2014.

\bibitem[Xiao et~al.(2021)Xiao, Dollar, Singh, Mintun, Darrell, and
  Girshick]{xiao2021early}
Xiao, T., Dollar, P., Singh, M., Mintun, E., Darrell, T., and Girshick, R.
\newblock Early convolutions help transformers see better.
\newblock \emph{Advances in Neural Information Processing Systems}, 34, 2021.

\bibitem[Xie et~al.(2017)Xie, Girshick, Doll{\'a}r, Tu, and
  He]{xie2017aggregated}
Xie, S., Girshick, R., Doll{\'a}r, P., Tu, Z., and He, K.
\newblock Aggregated residual transformations for deep neural networks.
\newblock In \emph{Proceedings of the IEEE conference on computer vision and
  pattern recognition}, pp.\  1492--1500, 2017.

\bibitem[Yen \& Finkel(1998)Yen and Finkel]{yen1998extraction}
Yen, S.-C. and Finkel, L.~H.
\newblock Extraction of perceptually salient contours by striate cortical
  networks.
\newblock \emph{Vision research}, 38\penalty0 (5):\penalty0 719--741, 1998.

\bibitem[Zoran et~al.(2020)Zoran, Chrzanowski, Huang, Gowal, Mott, and
  Kohli]{zoran2020towards}
Zoran, D., Chrzanowski, M., Huang, P.-S., Gowal, S., Mott, A., and Kohli, P.
\newblock Towards robust image classification using sequential attention
  models.
\newblock In \emph{Proceedings of the IEEE/CVF conference on computer vision
  and pattern recognition}, pp.\  9483--9492, 2020.

\bibitem[Zucker et~al.(1989)Zucker, Dobbins, and Iverson]{zucker1989two}
Zucker, S.~W., Dobbins, A., and Iverson, L.
\newblock Two stages of curve detection suggest two styles of visual
  computation.
\newblock \emph{Neural computation}, 1\penalty0 (1):\penalty0 68--81, 1989.

\end{thebibliography}
\bibliographystyle{icml2022}

\newpage
\appendix
\onecolumn

\section{Additional Qualitative Results}

\subsection{Evolution of Attention Maps during Recurrent Updates in \model}

\begin{figure}[h]
\vskip 0.2in
\begin{center}
\centerline{\includegraphics[width=0.5\columnwidth]{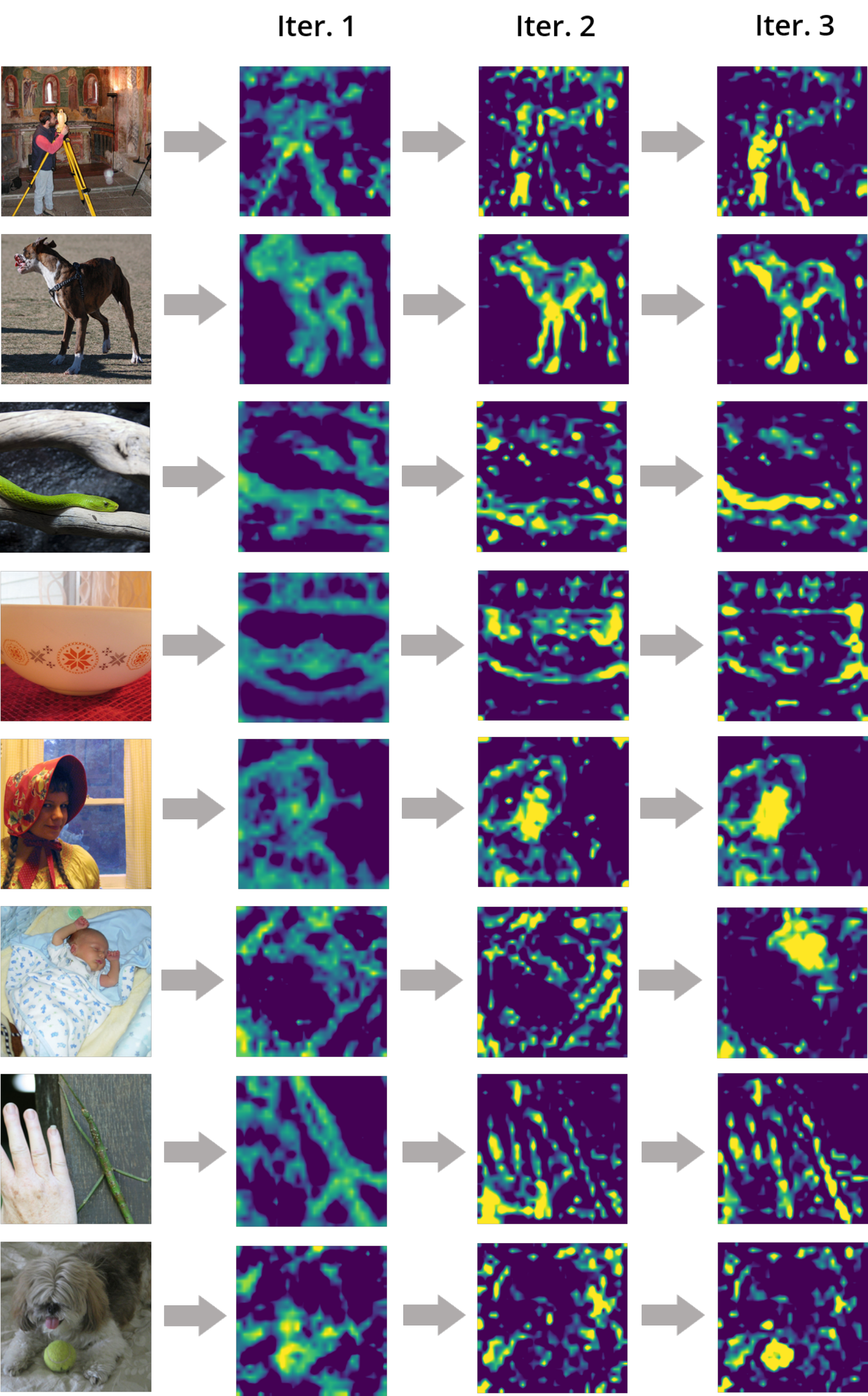}}
\caption{Visualization of the attention maps after each iteration of update in VARS. One can see that the attention will be more concentrated on the salient objects in the image after each update.}
\label{apdx:fig:att_each_iter}
\end{center}
\vskip -0.2in
\end{figure}

In Figure \ref{apdx:fig:att_each_iter} we show more examples of the evolution of attention maps in each step of the recurrent update in \model. Here we choose \model-D built upon the RVT baseline and visualize the attention map of the last layer in the first block. We can observe that the attention map is more sharp and concentrated on the salient objects after each iteration.

\vfill
\vfill

\newpage

\subsection{Visualization of Dynamic Dictionaries in \model-D and \model-SD}
\label{apdx:sec:dict_vis}

\begin{figure}[H]
\vskip 0.2in
\begin{center}
\centerline{\includegraphics[width=0.7\columnwidth]{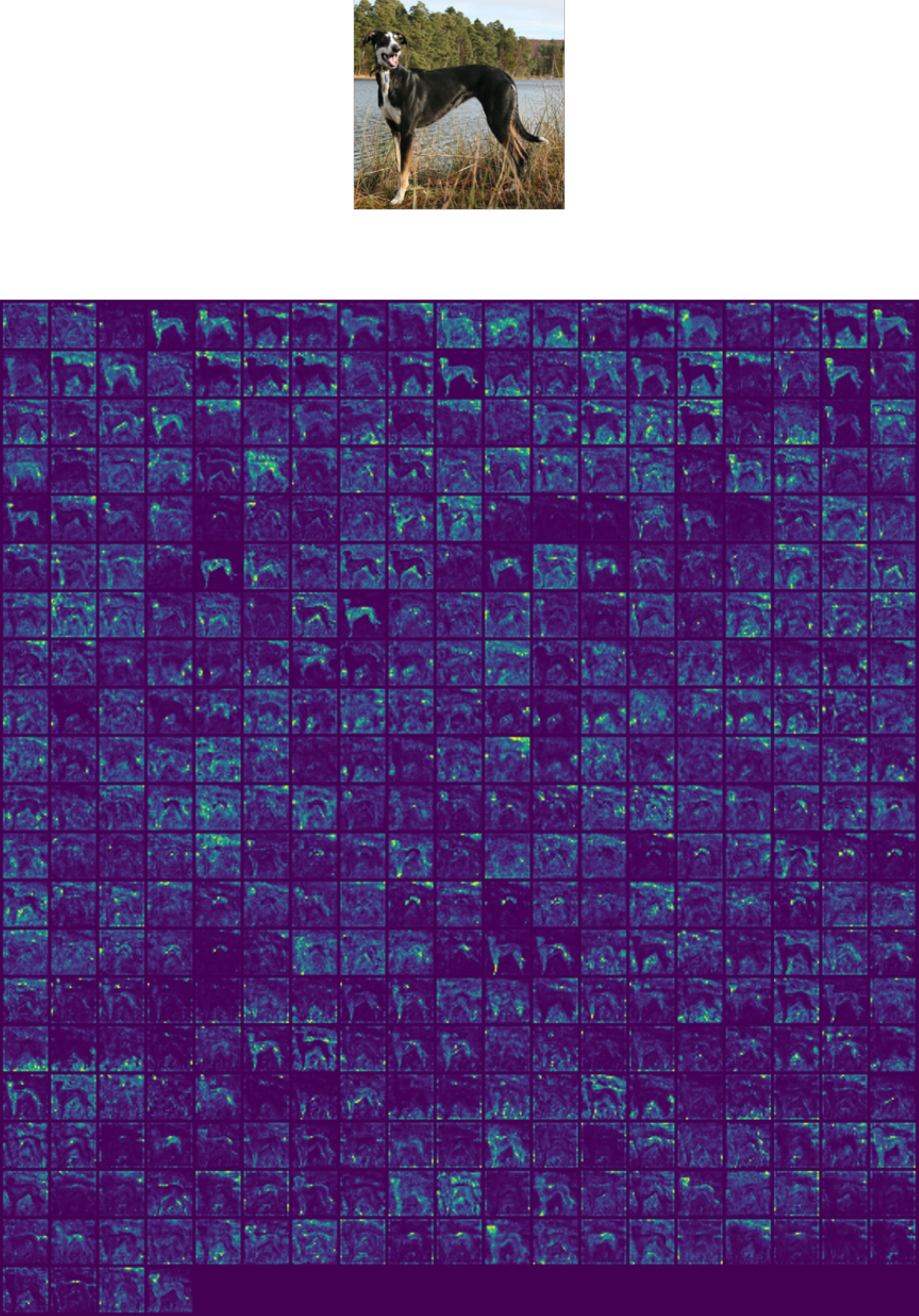}}
\caption{Visualization of the dynamic dictionary. The atoms are mostly masks on either foreground objects or backgrounds.}
\label{apdx:fig:dyn_dict_vis1}
\end{center}
\vskip -0.2in
\end{figure}

\begin{figure}[H]
\vskip 0.2in
\begin{center}
\centerline{\includegraphics[width=0.7\columnwidth]{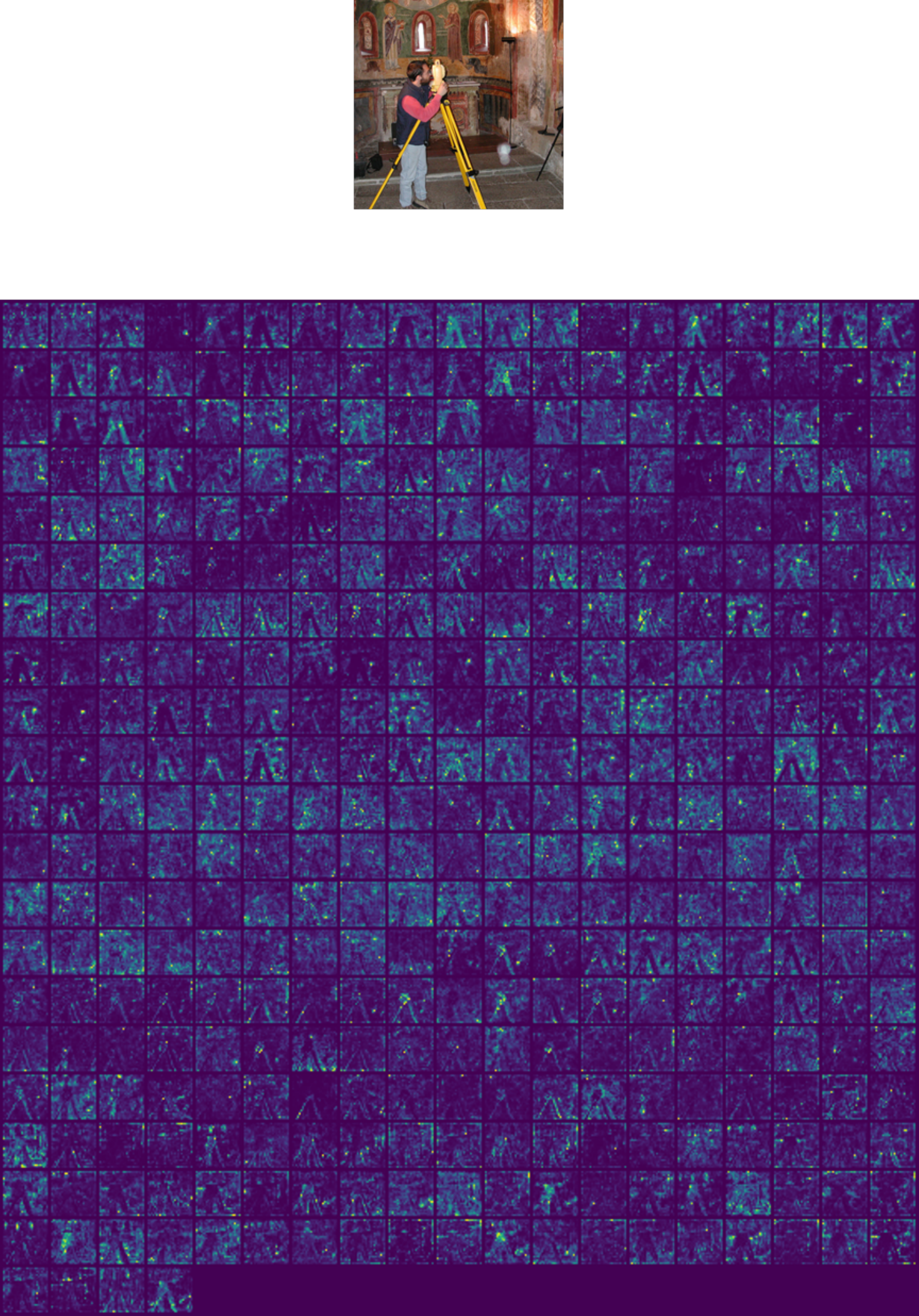}}
\caption{Visualization of the dynamic dictionary. The atoms are mostly masks on either foreground objects or backgrounds.}
\label{apdx:fig:dyn_dict_vis2}
\end{center}
\vskip -0.2in
\end{figure}

\begin{figure}[H]
\vskip 0.2in
\begin{center}
\centerline{\includegraphics[width=0.7\columnwidth]{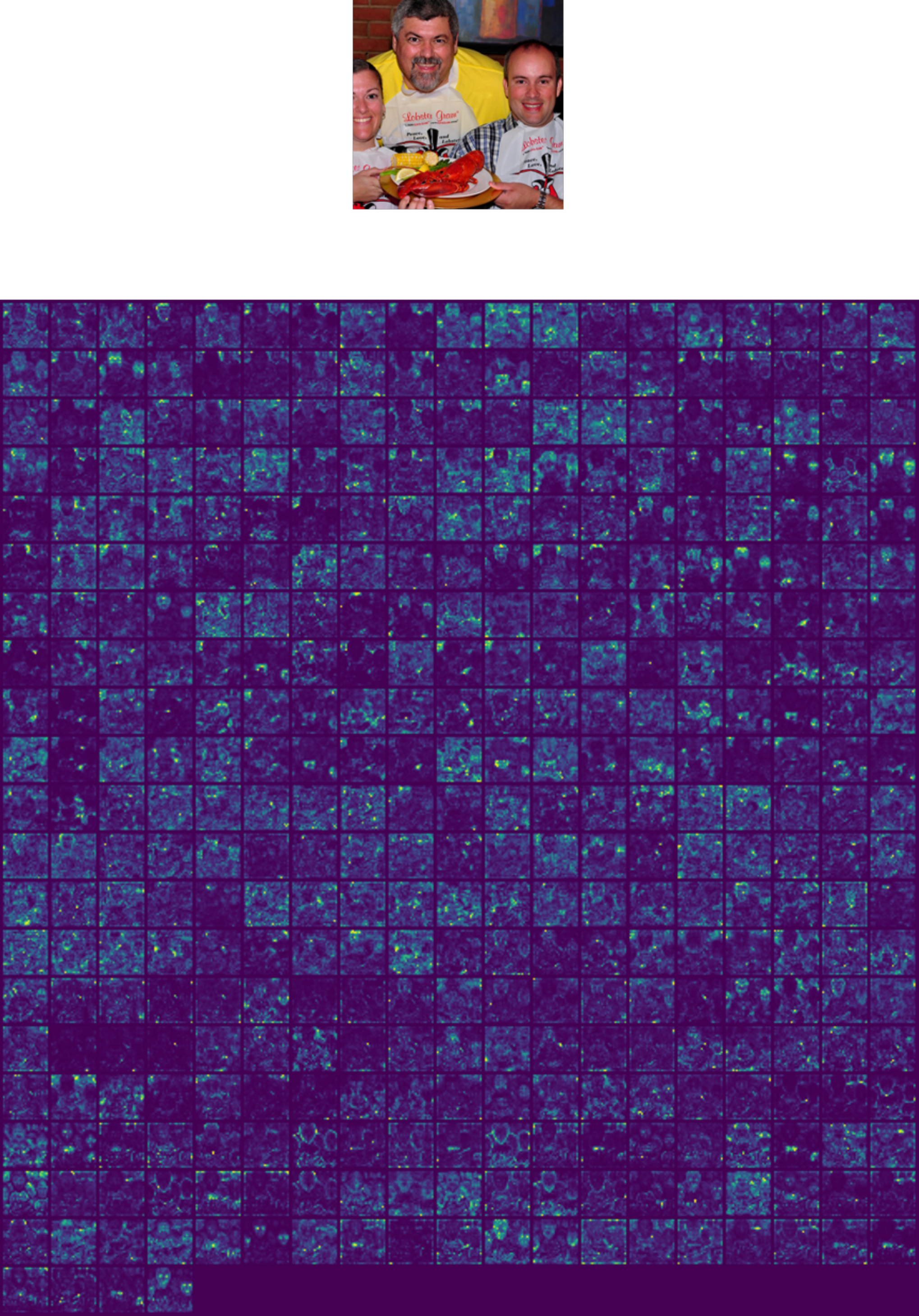}}
\caption{Visualization of the dynamic dictionary. The atoms are mostly masks on either foreground objects or backgrounds.}
\label{apdx:fig:dyn_dict_vis3}
\end{center}
\vskip -0.2in
\end{figure}

In \model-D and \model-SD, we use the input-dependent dictionary $\Phi(\mathbf{X})$ for sparse reconstruction. Here we visualize the dynamic dictionaries for a deeper understanding (Figure~\ref{apdx:fig:dyn_dict_vis1}-\ref{apdx:fig:dyn_dict_vis3}). We can see that most atoms in the dictionary are approximately uniform masks on either foreground or background regions. This is a direct consequence from the design of self-attention and random features~\cite{choromanski2020rethinking}.

\vfill
\vfill
\newpage

\vfill
\vfill
\newpage

\vfill
\vfill
\newpage

\subsection{Visualization of Attention Maps under Different Image Corruptions}

\begin{figure}[h]
\vskip 0.2in
\begin{center}
\centerline{\includegraphics[width=0.7\columnwidth]{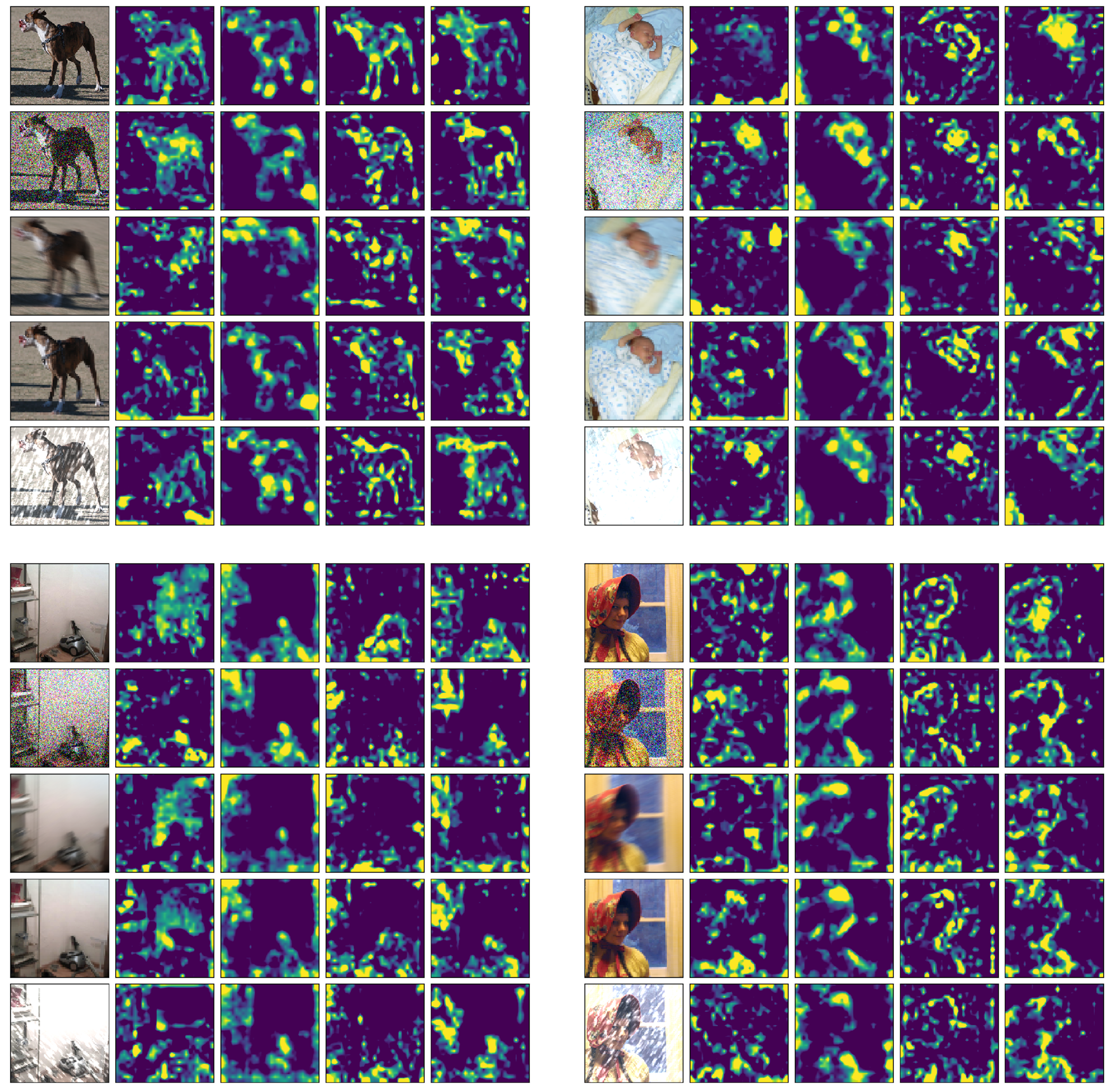}}
\caption{Visualization of the attention maps under image corruptions. Each block contains clean images as well as images under corruption of noise, blur, digital, and weather (from top to down). We visualize the attention map of RVT$^\ast$, VARS-S, VARS-D, and VARS-SD (from left to right in each block). Across different images, self-attention is usually more unstable than the variants of VARS. Meanwhile, VARS-S has attention maps that are consistent under different corruptions but are not as sharp as those of VARS-D and VARS-SD.}
\label{apdx:fig:att_noisy}
\end{center}
\vskip -0.2in
\end{figure}

We show additional visualization of the attention maps under different image corruptions in Figure \ref{apdx:fig:att_noisy}, where each block contains attention maps of clean images as well as images under noise, blur, digital, and weather corruptions (from top to down). We show the attention maps of RVT$^\ast$, VARS-S, VARS-D, and VARS-SD (from left to right). One can see that the attention maps of self-attention baseline is more sensitive to image corruptions, while variants of VARS tend to output stable attention maps. Meanwhile, the attention maps of VARS-S are steady but not as sharp as VARS-D and VARS-SD.

\vfill
\vfill

\newpage

\subsection{Comparing Attention Maps with Human Eye Fixation}

\begin{figure}[h]
\vskip 0.2in
\begin{center}
\centerline{\includegraphics[width=0.5\columnwidth]{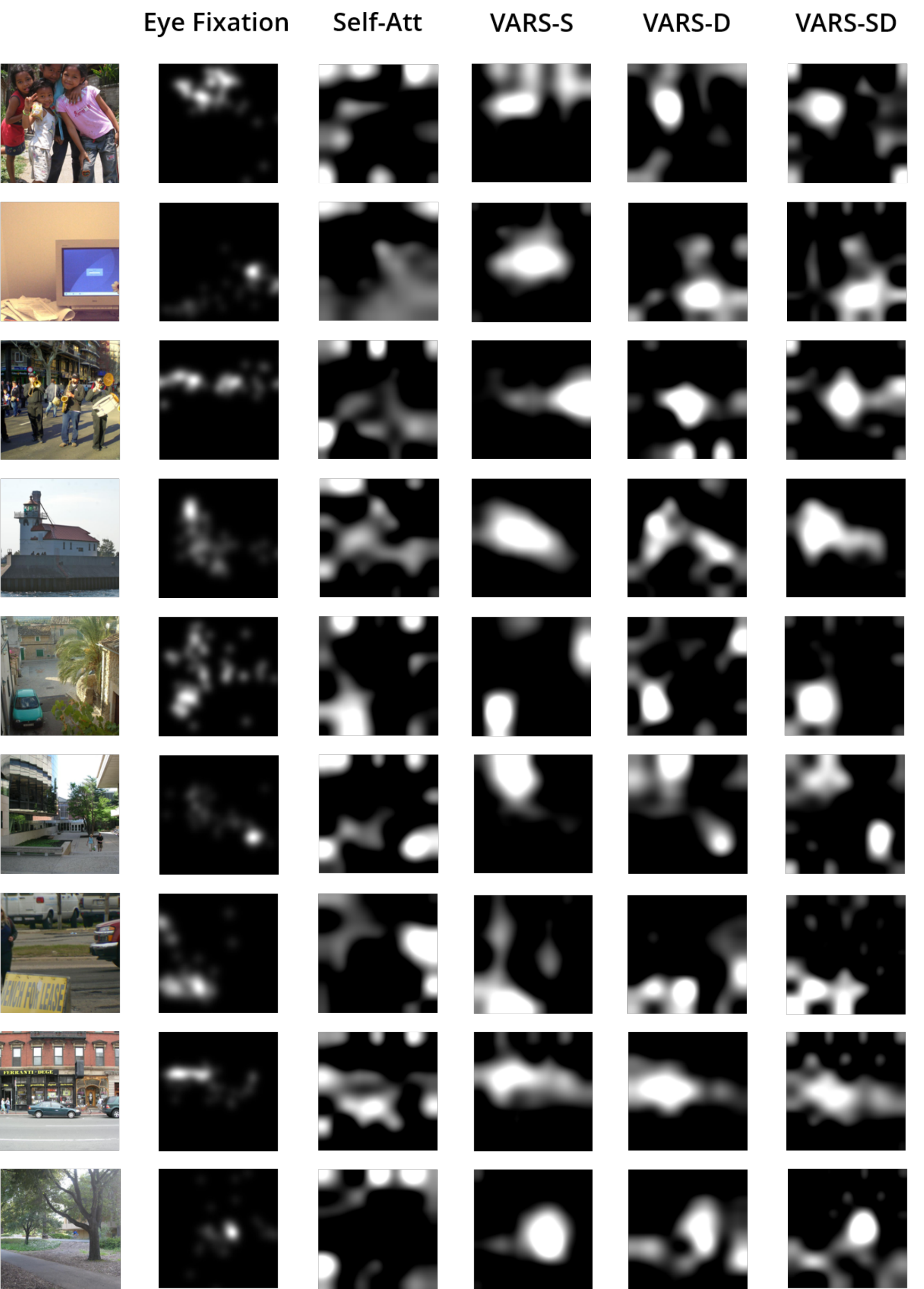}}
\caption{Comparison between attention maps of different models and human eye fixation probabilities. The variants of VARS have attention maps that are more consistent with human eye fixation.}
\label{apdx:fig:eye_fix}
\end{center}
\vskip -0.2in
\end{figure}

In Figure \ref{apdx:fig:eye_fix} we show addtional results on comparing the attention maps of different models with the human eye fixation data. We can see that, the attention maps of VARS are more consistent with human eye fixation than self-attention.

\vfill
\vfill

\newpage

\section{Details on Derivation of the Sparse Reconstruction Problem}

We follow~\citet{rozell2008sparse} and add recurrent connections between $\mathbf{u}$, modeled by the weight matrix $- \gamma (\mathbf{P}^T\mathbf{P} - \mathbf{I})$. We also add hyperparameters $\alpha$ and $\beta$ to
control the strength of self-leakage and the element-wise activation functions $\mathit{g}(\cdot)$ to gate the output from the neurons~\cite{dayan2001theoretical}. As a result, we fix the dynamics of the recurrent networks as:

\vspace{-0.5em}
\begin{small}
\begin{empheq}[left=\empheqlbrace]{align}
\label{apdx:eq:fixed_dynamics}
    &\dv{\mathbf{z}}{t} = -\alpha \mathbf{z} + \mathbf{P}\mathit{g}(\mathbf{u}) + \mathbf{x}, \\
    &\dv{\mathbf{u}}{t} = -\beta \mathbf{u} - \gamma (\mathbf{P}^T\mathbf{P} - \mathbf{I}) \mathit{g}(\mathbf{u}) + \mathbf{P}^T \mathbf{z} .
\end{empheq}
\end{small}\noindent
By taking $\alpha = 1$ and $\beta = \gamma = 2$, it has the same steady state solution as

\vspace{-0.5em}
\begin{small}
\begin{empheq}[left=\empheqlbrace]{align}
\label{apdx:eq:sparse_recon}
    \dv{\mathbf{u}}{t} &= -2(\mathbf{u} - \widetilde{\mathbf{u}}) -
    \mathbf{P}^T\mathbf{P} \widetilde{\mathbf{u}} + \mathbf{P}^T\mathbf{x}, \\
    \mathbf{z} &= \mathbf{P}\widetilde{\mathbf{u}} + \mathbf{x},
\end{empheq}
\end{small}\noindent
where $\widetilde{\mathbf{u}} = \mathit{g}(\mathbf{u})$. Now we choose $\mathit{g}(\cdot)$ as the thresholding function $\mathit{g}(\mathbf{u}_i) = \mathit{sgn}(\mathbf{u}_i) \cdot (\lvert \mathbf{u}_i \rvert - \lambda)_+$, where $\mathit{sgn}(\cdot)$ is the sign function and $(\cdot)_+$ is ReLU. Under the assumption that $\mathit{g}(\cdot)$ is monotonically non-decreasing, Eq.~\ref{apdx:eq:sparse_recon} is actually minimizing the energy function
\begin{equation}\small
\label{apdx:eq:lyapunov}
    E(\widetilde{\mathbf{u}}) = \frac{1}{2} ||\mathbf{P} \widetilde{\mathbf{u}} - \mathbf{x}||^2 + 2\lambda ||\widetilde{\mathbf{u}}||_1.
\end{equation}
To see this, one can verify that when $\mathbf{u}$ evolves by Eq.~\ref{apdx:eq:sparse_recon}, $E(\widetilde{\mathbf{u}})$ is non-increasing, \ie,
\begin{equation}\small
\label{apdx:eq:energy_evolution}
    \dv{E}{t} = -(2\lambda \cdot \mathit{sgn}(\mathbf{u})  + 
    \mathbf{P}^T\mathbf{P} \widetilde{\mathbf{u}} - \mathbf{P}^T\mathbf{x})^T \cdot \mathbf{K}(\mathbf{u}) \cdot (2\lambda \cdot \mathit{sgn}(\mathbf{u})  +
    \mathbf{P}^T\mathbf{P} \widetilde{\mathbf{u}} - \mathbf{P}^T\mathbf{x}) = (\dv{\mathbf{u}}{t})^T \mathbf{K}(\mathbf{u}) \dv{\mathbf{u}}{t},
\end{equation}
where $\mathbf{K}(\mathbf{u})$ is a diagonal matrix with $\mathbf{K}(\mathbf{u})_{ii} = 1$ when \emph{(i)} $|\mathbf{u}_i| > 1$, and \emph{(ii)} $|\mathbf{u}_i| = 1$ and $\dv{\mathbf{u}_i}{t} \cdot \mathit{sgn}(\mathbf{u}_i) > 0$, otherwise $\mathbf{K}(\mathbf{u})_{ii} = 0$. Since $\mathbf{K}(\mathbf{u})$ is positive semi-definite, Eq.~\ref{apdx:eq:energy_evolution} is non-positive, which means the energy is non-increasing. Then Eq.~\ref{apdx:eq:sparse_recon} equivalently optimizes the sparse reconstruction:

\vspace{-1em}
\begin{small}
\begin{empheq}[left=\empheqlbrace]{align}
\label{apdx:eq:static_a}
    \widetilde{\mathbf{u}}^\ast &= \argmin_{\widetilde{\mathbf{u}} \in \mathbb{R}^{d^\prime}} \frac{1}{2} ||\mathbf{P} \widetilde{\mathbf{u}} - \mathbf{x}||^2 + 2\lambda ||\widetilde{\mathbf{u}}||_1 \\
\label{apdx:eq:static_b}
    \mathbf{z}^\ast &= \mathbf{P}\widetilde{\mathbf{u}}^\ast + \mathbf{x}.
\end{empheq}
\end{small}

\end{document}